\documentclass[10pt,twocolumn,letterpaper]{article}

\usepackage{cvpr}              

\usepackage[accsupp]{axessibility}  
\usepackage{graphicx}
\usepackage{amsmath}
\usepackage{amssymb}
\usepackage{booktabs}

\usepackage{bm}
\usepackage{xcolor}
\usepackage{color, colortbl}
\usepackage{algorithmic}
\usepackage[linesnumbered,ruled,vlined]{algorithm2e}
\usepackage{threeparttable}
\usepackage{multirow}
\usepackage{bbding}

\usepackage[pagebackref,breaklinks,colorlinks]{hyperref}

\usepackage[capitalize]{cleveref}
\crefname{section}{Sec.}{Secs.}
\Crefname{section}{Section}{Sections}
\Crefname{table}{Table}{Tables}
\crefname{table}{Tab.}{Tabs.}

\def\cvprPaperID{9689} 
\def\confName{CVPR}
\def\confYear{2023}

\newcommand{\BU}[1]{#1}

\newcommand{\BV}[1]{\color{black!10!blue}{\em #1}}
\newcommand{\redbold}[1]{{\textbf{#1}}}

\newcommand{\setting}{PTTA\xspace}
\newcommand{\method}{RoTTA\xspace}
\newcommand{\source}{\mathcal{S}}

\newcommand{\distribution}{\mathcal{P}}
\newcommand{\stream}{\mathcal{X}}
\newcommand{\R}{\mathbb{R}}
\newcommand{\bank}{\mathcal{M}}
\newcommand{\capacity}{\mathcal{N}}
\newcommand{\age}{\mathcal{A}}
\newcommand{\uncertainty}{\mathcal{U}}
\newcommand{\numclass}{\mathcal{C}}
\newcommand{\sampling}{CSTU\xspace}
\newcommand{\loss}{\mathcal{L}}

\newcommand{\argmax}{\mathop{\arg\max}}

\renewcommand{\paragraph}[1]{\vspace{1.25mm}\noindent\textbf{#1}}

\newcommand{\VspaceBefore}{\vspace{-3mm}}
\newcommand{\VspaceAfter}{\vspace{-3mm}}

\definecolor{Gray}{gray}{0.9}

\definecolor{text01purple}{RGB}{168,119,200}
\newcommand{\textpurple}[1]{\textcolor{text01purple}{#1}}

\definecolor{text01green}{RGB}{82,208,83}
\newcommand{\textgreen}[1]{\textcolor{text01green}{#1}}

\definecolor{text02red}{RGB}{211,41,15}
\newcommand{\textred}[1]{\textcolor{text02red}{#1}}

\definecolor{text02yellow}{RGB}{230,119,11}
\newcommand{\textyellow}[1]{\textcolor{text02yellow}{#1}}

\newcommand{\dtplus}[1]{\fontsize{6pt}{0.1em}\selectfont (\textbf{\textgreen{#1}})}

\newcommand{\gain}[1]{\textgreen{#1}}

\makeatletter
\renewcommand*{\@fnsymbol}[1]{\ensuremath{\ifcase#1\or \text{\Envelope}\or \dagger\or \ddagger\or
   \mathsection\or \mathparagraph\or \|\or **\or \dagger\dagger
   \or \ddagger\ddagger \else\@ctrerr\fi}}
\makeatother

\begin{document}

\title{Robust Test-Time Adaptation in Dynamic Scenarios}

\author{Longhui Yuan \quad Binhui Xie \quad Shuang Li\thanks{Corresponding author} \\
School of Computer Science and Technology, Beijing Institute of Technology \\
{\tt\small \{longhuiyuan,binhuixie,shuangli\}@bit.edu.cn}}

\maketitle

\begin{abstract}
    Test-time adaptation (TTA) intends to adapt the pre-trained model to test distributions with only unlabeled test data streams. 
    Most of the previous TTA methods have achieved great success on simple test data streams such as independently sampled data from single or multiple distributions. 
    However, these attempts may fail in dynamic scenarios of real-world applications like autonomous driving, where the environments gradually change and the test data is sampled correlatively over time. 
    In this work, we explore such practical test data streams to deploy the model on the fly, namely practical test-time adaptation (\setting). 
    To do so, we elaborate a {\bf Ro}bust {\bf T}est-{\bf T}ime {\bf A}daptation (\method) method against the complex data stream in PTTA. 
    More specifically, we present a robust batch normalization scheme to estimate the normalization statistics. 
    Meanwhile, a memory bank is utilized to sample category-balanced data with consideration of timeliness and uncertainty. 
    Further, to stabilize the training procedure, we develop a time-aware reweighting strategy with a teacher-student model. 
    Extensive experiments prove that \method enables continual test-time adaptation on the correlatively sampled data streams.
    Our method is easy to implement, making it a good choice for rapid deployment. 
    The code is publicly available at \url{https://github.com/BIT-DA/RoTTA}

\end{abstract}

\section{Introduction}
\label{sec:intro}

In recent years, many machine learning problems have made considerable headway with the success of deep neural networks~\cite{DLeCunBH15,alexnet,resnet,DosovitskiyB0WZ21}. Unfortunately, the performance of deep models drops significantly when training data and testing data come from different distributions~\cite{quinonero2008dataset}, which limits their utility in real-world applications. To reduce the distribution shift, a handful of works focus on transfer learning field~\cite{survey}, in particular, domain adaptation (DA)~\cite{wang2018deep,GaninUAGLLML16,DAN-PAMI,GDCAN,Tzeng_ADDA,LiangHF20} or domain generalization (DG)~\cite{zhou2022domain,wang2022generalizing,MuandetBS13,LiPWK18,LiYSH18}, in which one or more different but related labeled datasets (a.k.a. source domain) are collected to help the model generalize well to unlabeled or unseen samples in new datasets (a.k.a. target domain). 

\begin{figure}[t]
    \includegraphics[width=\linewidth]{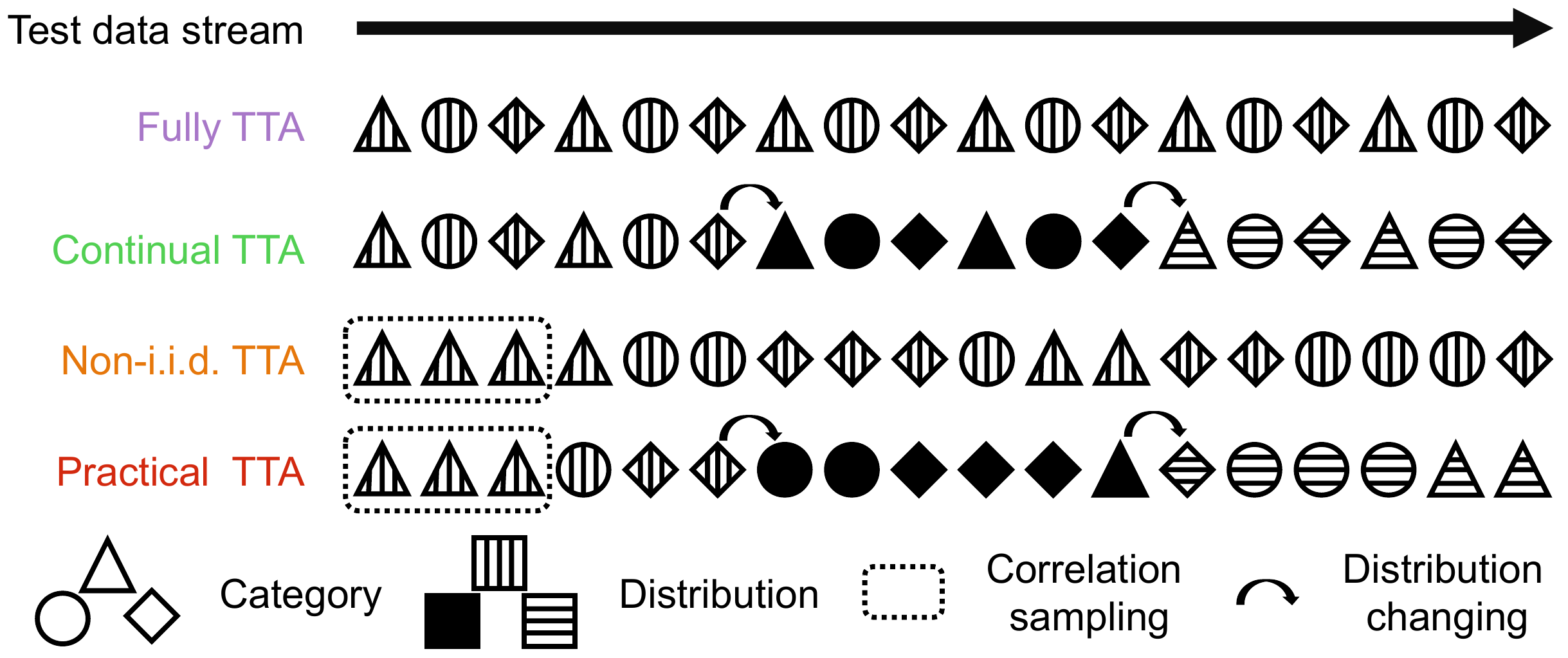}
    \vspace{-6mm}
    \caption{We consider the practical test-time adaptation (TTA) setup and compare it with related ones. First, \textpurple{Fully TTA}~\cite{tent_wang2020} adapts models on a fixed test distribution with an independently sampled test stream. Then, on this basis, \textgreen{Continual TTA}~\cite{cotta} takes the continually changing distributions into consideration. Next, \textyellow{Non-i.i.d. TTA}~\cite{note} tries to tackle the correlatively sampled test streams on a single test distribution, where the label distribution among a batch of data deviates from that of the test distribution. To be more practical, \textred{Practical TTA} strives to connect both worlds: distribution changing and correlation sampling.}
    \label{fig:settings}
    \vspace{-4mm}
\end{figure}

While both DA and DG have extensively studied the problem of distribution shifts, they typically assume accessibility to the raw source data. However, in many practical scenarios like personal consumption records, the raw data should not be publicly available due to data protection regulations. Further, existing methods have to perform heavy backward computation, resulting in unbearable training costs. Test-time adaptation (TTA)~\cite{ttt_sun2020,ChoiYCY22,NiuW0CZZT22,BatesonLA22,huang2022extrapolative,IwasawaM21,gandelsman2022testtime,zhang2022memo} attempts to address the distribution shift online at test time with only unlabeled test data streams. Unequivocally, TTA has drawn widespread attention in a variety of applications, e.g., 2D/3D visual recognition~\cite{ttt_sun2020,MaCZQZD22,KimH022,AzimiPRHB022,ZhangBCP22}, multimodality~\cite{shu2022testtime,ShinTZSLGKY22} and document understanding~\cite{tta_document}.

\begin{table*}[t]
    \centering
    \caption{Comparison between our proposed practical test-time adaptation (PTTA) and related adaptation settings. 
    }
    \label{table:settings}
    \VspaceBefore
    \resizebox{0.86\textwidth}{!}{
    \begin{tabular}{l | c c | c c | c c}
    \toprule[1.2pt]
     \multirow{2}{*}{Setting} & \multicolumn{2}{c |}{Adaptation Stage} & \multicolumn{2}{c |}{Available Data} & \multicolumn{2}{c}{Test Data Stream} \\
    
    \cline{2-7}
    
     & Train & Test & Source & Target & Distribution & Sampling Protocol \\

    \hline

    Domain Adaptation & \Checkmark & \XSolidBrush & \Checkmark & \Checkmark & - & - \\

    Domain Generalization & \Checkmark & \XSolidBrush & \Checkmark & \XSolidBrush & - & - \\
    
    Test-Time Training~\cite{ttt_sun2020} & \Checkmark & \Checkmark & \Checkmark & \Checkmark & stationary & independently \\

    Fully Test-Time Adaptation~\cite{tent_wang2020} & \XSolidBrush & \Checkmark & \XSolidBrush & \Checkmark & stationary & independently \\

    Continual Test-Time Adaptation~\cite{cotta}  & \XSolidBrush & \Checkmark & \XSolidBrush & \Checkmark & continually changing & independently \\

    Non-i.i.d. Test-Time Adaptation~\cite{niid_boudiaf2022parameter,note}  & \XSolidBrush & \Checkmark & \XSolidBrush & \Checkmark & stationary & correlatively \\

    \hline

    \bf Practical Test-Time Adaptation (Ours) & \XSolidBrush & \Checkmark & \XSolidBrush & \Checkmark & continually changing & correlatively \\
    
    \bottomrule[1.2pt]
    \end{tabular}
    }
    \vspace{-4mm}
\end{table*}

Prior TTA studies~\cite{tent_wang2020,chen2022_contrastivetta,goyal2022test,cotta} mostly concentrate on a simple adaptation scenario, where test samples are independently sampled from a fixed target domain. To name a few, Sun \etal~\cite{ttt_sun2020} adapt to online test samples drawn from a constant or smoothly changing distribution with an auxiliary self-supervised task. Wang \etal~\cite{tent_wang2020} adapt to a fixed target distribution by performing entropy minimization online.
However, such an assumption is violated when the test environments change frequently~\cite{cotta}. 
Later on, Boudiaf \etal~\cite{niid_boudiaf2022parameter} and Gong \etal~\cite{note} consider the temporal correlation ship within test samples. 
For example, in autonomous driving, test samples are highly correlated over time as the car will follow more vehicles on the highway or will encounter more pedestrians in the streets. More realistically, the data distribution changes as the surrounding environment alerts in weather, location, or other factors. 
In a word, distribution change and data correlation occur simultaneously in reality. 

Confronting continually changing distributions, traditional algorithms like pseudo labeling or entropy minimization become more unreliable as the error gradients cumulate. Moreover, the high correlation among test samples results in the erroneous estimation of statistics for batch normalization and collapse of the model. 
Driven by this analysis, adapting to such data streams will encounter two major obstacles: 1) incorrect estimation in the batch normalization statistics leads to erroneous predictions of test samples, consequently resulting in invalid adaptation; 2) the model will easily or quickly overfit to the distribution caused by the correlative sampling. 
Thus, such dynamic scenarios are pressing for a new TTA paradigm to realize robust adaptation. 

In this work, we launch a more realistic TTA setting, where distribution changing and correlative sampling occur simultaneously at the test phase. We call this \textit{Practical Test-Time Adaptation}, or briefly, \textit{PTTA}. To understand more clearly the similarities and differences between PTTA and the previous setups, we visualize them in Figure~\ref{fig:settings} and summarize them in Table~\ref{table:settings}. To conquer this challenging problem, we propose a {\bf Ro}bust {\bf T}est-{\bf T}ime {\bf A}daptation ({\bf \method}) method, which consists of three parts: 1) robust statistics estimation, 2) category-balanced sampling considering timeliness and uncertainty and 3) time-aware robust training. More concretely, we first replace the erroneous statistics of the current batch with global ones maintained by the exponential moving average. It is a more stable manner to estimate the statistics in BatchNorm layers.
Then, we simulate a batch of independent-like data in memory with category-balanced sampling while considering the timeliness and uncertainty of the buffered samples. That is, samples that are newer and less uncertain are kept in memory with higher priority. With this batch of category-balanced, timely and confident samples, we can obtain a snapshot of the current distribution. Finally, we introduce a time-aware reweighting strategy that considers the timeliness of the samples in the memory bank, with a teacher-student model to perform robust adaptation. With extensive experiments, we demonstrate that \method can robustly adapt in the practical setup, i.e., PTTA.

In a nutshell, our contributions can be summarized as:\vspace{-2mm}
\begin{itemize}
    \item We propose a new test-time adaptation setup that is more suitable for real-world applications, namely practical test-time adaptation (PTTA). PTTA considers both distribution changing and correlation sampling.\vspace{-2mm}
    \item We benchmark the performance of prior methods in PTTA and uncover that they only consider one aspect of the problem, resulting in ineffective adaptation.\vspace{-2mm}
    \item We propose a robust test-time adaptation method (\method), which has a more comprehensive consideration of PTTA challenges. Ease of implementation and effectiveness make it a practical deployment option. \vspace{-2mm}
    \item We extensively demonstrate the practicality of PTTA and the effectiveness of \method on common TTA benchmarks~\cite{corruptions}, i.e., CIFAR-10-C and CIFAR-100-C and a large-scale DomainNet~\cite{DomainNet} dataset. \method obtains state-of-the-art results, outperforming the best baseline by a large margin (reducing the averaged classification error by over 5.9\%, 5.5\% and 2.2\% on CIFAR-10-C, CIFAR-100-C and DomainNet, respectively).
\end{itemize}

\begin{figure*}[t]\centering
    \includegraphics[width=0.97\linewidth]{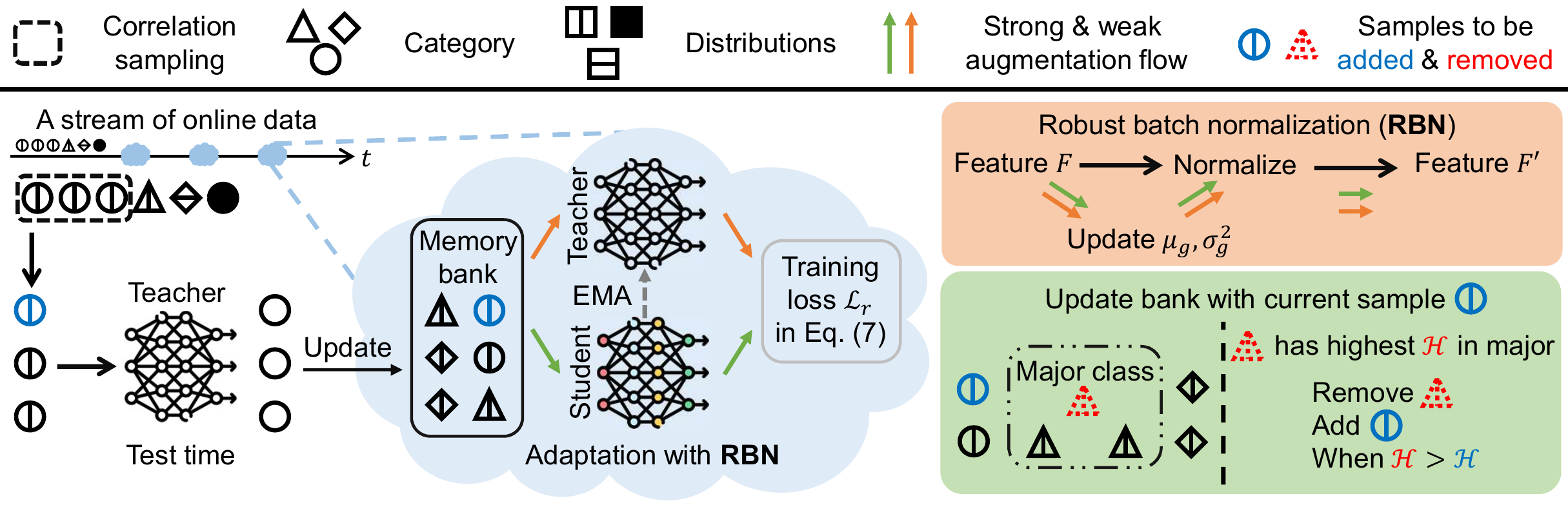}
    \VspaceAfter
    \caption{{\bf Framework overview.} Firstly, we replace the batch normalization layer with RBN which robustly normalizes the feature map. During the inference of the online test stream of PTTA, we utilize the predictions of samples to maintain a memory bank by category-balanced sampling with timeliness and uncertainty. Finally, we use the category-balanced, timely and confident data in the memory bank combined with a robust loss to adapt the model at test time.}
    \VspaceAfter
    \label{fig:framework}
\end{figure*}

\section{Related Work}
\label{sec:related}
\paragraph{Domain adaptation (DA)} studies the problem of transferring the knowledge learned from a labeled source dataset to an unlabeled target dataset~\cite{MansourMR09,GaninUAGLLML16,Tzeng_2015_Simultaneous,tsai2018learning,chen2018domain,TSA}. Representative techniques include latent distribution alignment~\cite{DAN-PAMI,AFN2019Xu}, adversarial training~\cite{GaninUAGLLML16,MCD}, or self-training~\cite{zou2018unsupervised,sepico}. 
The limitation of this setting, however, is that an unlabeled test dataset (target domain) is needed at training time, in addition to a labeled training dataset (source domain). 
Accordingly, it might fail to handle more practical scenarios like test-time adaptation. 
Our practical test-time adaptation setting can be viewed as performing correlatively sample adaptation on the fly. It is worth noting that standard domain adaptation techniques might collapse when only continual data streams from multiple target domains are accessible.

\paragraph{Domain generalization (DG)} assumes that multiple source domains are available for model training and tries to learn models that can generalize well to any unseen domains~\cite{BlanchardLS11,MuandetBS13,LiPWK18,LiYSH18,ZhouY0X21,IwasawaM21}. A broad spectrum of methodologies based on data augmentation~\cite{ZhouY0X21,XuLYRN21}, meta-learning~\cite{LiYSH18,DuXXQZS020}, or domain alignment~\cite{MahajanTS21,MuandetBS13} has made great progress. In contrast, this work instead aims to improve the performance of source pre-trained models at the test time by using unlabeled online data streams from multiple continually changing target domains.

\paragraph{Continual learning (CL)} (also known as incremental learning, life-long learning) addresses the problem of learning a model for many tasks sequentially without forgetting knowledge obtained from the preceding tasks.~\cite{KirkpatrickPRVD16,AljundiLGB19,LangeAMPJLST22,castro2018end,RebuffiKSL17}. CL methods can often be categorized into replay-based~\cite{RebuffiKSL17,TiwariKIS22} and regularization-based~\cite{KirkpatrickPRVD16,LiH18a} methods.
Ideas from continual learning are also adopted for continuous domain adaptation approaches~\cite{wulfmeier2018incremental,kumar2020understanding}
In our work, we share the same motivation as CL and point out that practical test-time adaptation (PTTA) also suffers catastrophic forgetting (i.e., performance degradation on new test samples due to correlation sampling), which makes test-time adaptation approaches are unstable to deploy. 

\paragraph{Test-time adaptation (TTA)} focus on more challenging settings where only source model and unlabeled target data are available~\cite{jain2011online,tta_iclr1,LiuKDBMA21,ChiWYT21,vpt_tta,KunduVVB20,royer2015classifier}. A similar paradigm is source-free domain adaptation (SFDA)~\cite{ChidlovskiiCC16,yang2021generalized,liu2021source,kurmi2021domain}, which also requires no access to the training (source) data.
To name a few, Liang \etal~\cite{LiangHF20} fit the source hypothesis by exploiting the information maximization and self-supervised pseudo-labeling. Kundu \etal~\cite{KunduVVB20} formalize a unified solution that explores SFDA without any category-gap knowledge.
To fully utilize any arbitrary pre-trained model, Sun \etal~\cite{ttt_sun2020} propose conducting adaptation on the fly with an auxiliary self-supervised task. Later on, Wang \etal~\cite{tent_wang2020} take a source pre-trained model and adapt it to the test data by updating a few trainable parameters in BatchNorm layers~\cite{BN} using entropy minimization~\cite{entropy_minimization}. 

While standard TTA has been widely studied in many tasks~\cite{tent_wang2020,goyal2022test,shu2022testtime,ShinTZSLGKY22,ZhangBCP22,AzimiPRHB022}, the fact remains that both distribution changing~\cite{cotta} and data correlation sampling~\cite{note} has only been considered in isolation.
For example, Gong \etal~\cite{note} propose instance-aware batch normalization and prediction-balanced reservoir sampling to address the challenges of correlatively sampled test streams, however, it does not consider unstable adaptation resulting from long-term adaptation on continually changing distributions. On the other hand, Wang \etal~\cite{cotta} assume that the target test data is streamed from a continually changing environment and continually adapt an off-the-shelf source pre-trained model to the current test data. In this work, we launch PTTA, a more practical TTA setting to connect both worlds: distribution changing and correlation sampling.

\section{Method}\label{sec:method}

\begin{figure}[t]\centering
    \includegraphics[width=0.94\linewidth]{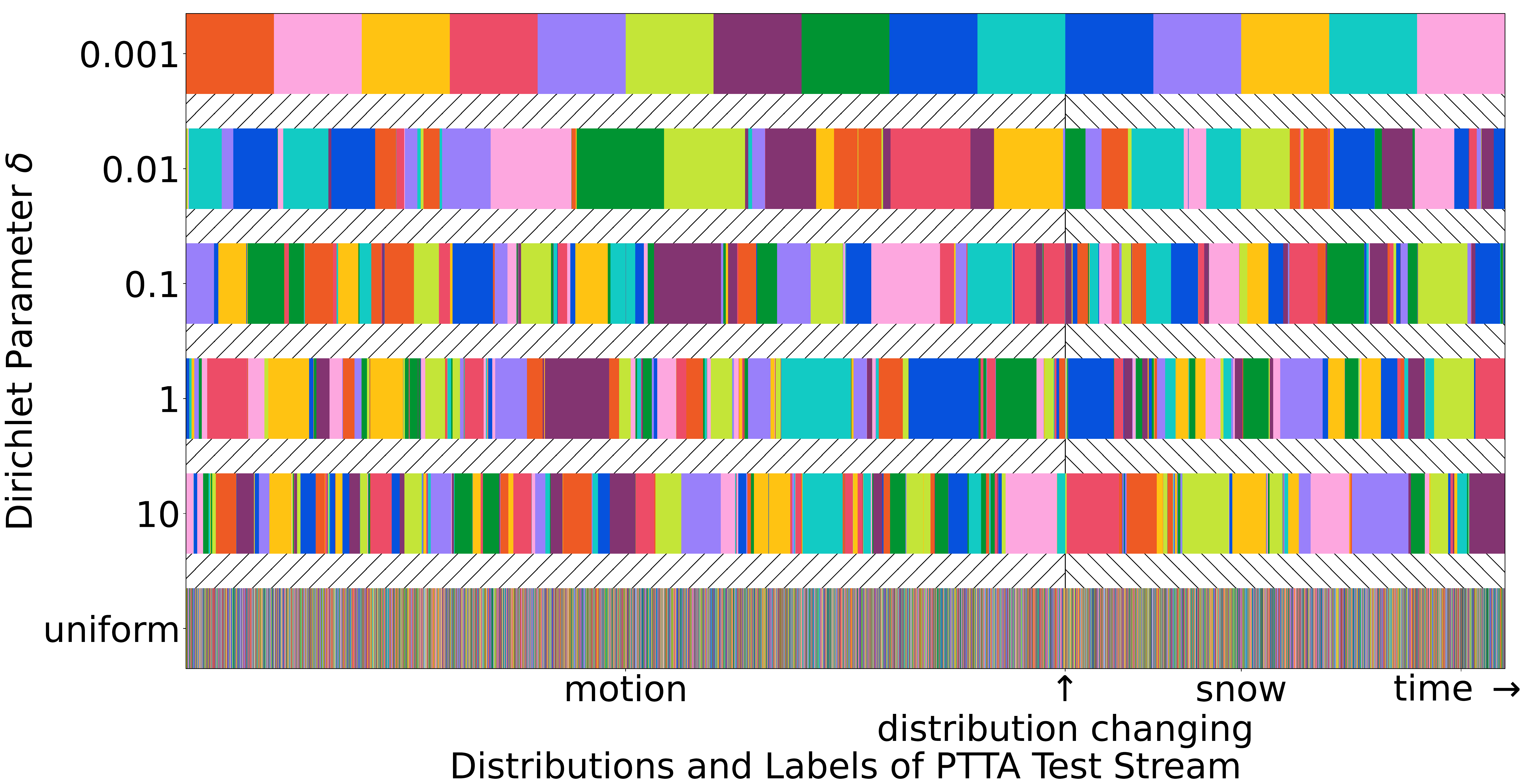}
    \VspaceBefore
    \caption{Illustration of the labels and distributions of the test stream of CIFAR10-C under the setup \setting. And we adopt Dirichlet distribution to simulate the process of correlative sampling. It is clear that as the concentration parameter $\delta$ decreases, the correlation among sampled data increases, which is reflected in the increasing aggregation of categories.}
    \VspaceAfter
    \label{fig:test_stream}
\end{figure}

\subsection{Problem Definition and Motivation}
\label{sec:problem_ptta}
Given a model $f_{\theta_0}$ with parameter $\theta_0$ pre-trained on source domain $\mathcal{D_\source}=\{(x^\source, y^\source)\}$, the proposed practical test-time adaptation (PTTA) aims to adapt $f_{\theta_0}$ to a stream of online unlabeled samples $\stream_0, \stream_1, ..., \stream_T$, where $\stream_t$ is a batch of highly correlated samples from the distribution $\distribution_{test}$ that changes with time $t$ continually. More specifically, at test time, with time going on, the test distribution $\distribution_{test}$ changes continually as $\distribution_0, \distribution_1, ...,\distribution_\infty$. At time step $t$, we will receive a batch of unlabeled and correlated samples $\stream_t$ from $\distribution_{test}$. Next, $\stream_t$ is fed into the model $f_{\theta_t}$ and the model needs to adapt itself to the current test data streams and make predictions $f_{\theta_t}(\stream_t)$ on the fly. 

As a matter of fact, this setup is largely driven the practical demands of deploying models in dynamic scenarios. Taking for example the case of autonomous driving mentioned in \S~\ref{sec:intro}, test samples are highly correlated and the data distribution changes continually with the weather or location. Another example is the situation of intelligent monitoring, the camera will continuously capture more people at certain times, such as after work, but fewer of them during work time. Meanwhile, the light condition changes continually from day to night. The deployed model should be robustly adapted in such dynamic scenarios. In a word, distribution change and data correlation often happen simultaneously in the real world. For this reason, existing TTA methods~\cite{tta_iclr1,tent_wang2020,zhang2022memo,ChiWYT21,chen2022_contrastivetta,cotta,note} might become unstable when the test stream is sampled from such dynamic scenarios. 

To obtain the test stream of \setting, we adopt Dirichlet Distribution with parameter $\delta$ to simulate the correlation among test samples. We present the test data streams corresponding to different values of $\delta$ on the CIFAR10-C dataset in Figure~\ref{fig:test_stream}. We can observe that the smaller $\delta$ is, the higher the correlation will be. For the sake of unity, we set $\delta = 0.1$ as the default for all experiments. In the following, we present a robust test-time adaptation framework for the practical test-time adaptation setup defined above. An overview of our \method is illustrated in Figure~\ref{fig:framework}. 

\subsection{Robust Test-Time Adaptation}
\label{sec:robust_tta}

Motivated by the fact that the statistics of current batch data, which are commonly used in previous TTA methods~\cite{tent_wang2020,chen2022_contrastivetta,goyal2022test,cotta,ttt_sun2020}, become unreliable when they encounter correlative test data streams, we first turn to the global robust statistics for normalization.
Then, to effectively adapt to the current distribution, we maintain a memory bank by category-balanced sampling with considering timeliness and uncertainty, which captures a more stable snapshot of the distribution. 
Finally, we utilize the teacher-student model and design a timeliness-based reweighting strategy to train the model robustly. 

\paragraph{Robust batch normalization (RBN).}
\label{sec:RBN}
Batch Normalization (BN)~\cite{BN} is a widely-used training technique as it can accelerate the training and convergence speed of networks and stabilize the training process by reducing the risk of gradient explosion and vanishing.
Given the feature map $F \in \R^{B\times C\times H\times W}$ as the input for a BN layer when training, the channel-wise mean $\mu \in \R^C$ and variance $\sigma^2 \in \R^C$ are calculated as follows:
\begin{small}
    \begin{align}
        \label{eq:mu_b}
        \mu_c &= \frac{1}{BHW}\sum_{b=1}^B \sum_{h=1}^H \sum_{w=1}^W F_{(b,c,h,w)}\,, \\
        \label{eq:sigma_b}
        \sigma^2_c &= \frac{1}{BHW}\sum_{b=1}^B \sum_{h=1}^H \sum_{w=1}^W (F_{(b,c,h,w)} - \mu_c)^2\,.
    \end{align}
\end{small}%
Then the feature map is normalized and refined in a channel-wise manner as 
\begin{small}
\begin{align}
    BN(F_{(b,c,h,w)};\mu,\sigma^2) = \gamma_c\frac{F_{(b,c,h,w)}-\mu_c}{\sqrt{\sigma^2_c+\epsilon}} + \beta_c\,,\label{eq:affine}
\end{align}%
\end{small}%
where $\gamma,\beta\in\R^C$ are learnable parameters in the layer and $\epsilon>0$ is a constant for numerical stability. 
Meanwhile, during training, the BN layer maintains a group of global running mean and running variance $(\mu_s, \sigma^2_s)$ for inference.

Due to the domain shift at test time, the global statistics $(\mu_s, \sigma^2_s)$ normalize test features inaccurately, causing significant performance degradation.
To tackle the problem above, some methods~\cite{tent_wang2020,cotta,EATA} use the statistics of the current batch to perform normalization.
Unfortunately, when the test samples have a high correlation under \setting setup, the statistics of the current batch also fail to correctly normalize the feature map, as demonstrated in Figure~\ref{fig:analyze_delta}. Specifically, the performance of BN~\cite{BN_Stat} decreases rapidly as the data correlation increases.

Based on the analysis above, we propose a robust batch normalization (RBN) module, which maintains a group of global statistics $(\mu_g,\sigma_g^2)$ to normalize the feature map robustly.
Before the whole test-time adaptation, $(\mu_g,\sigma_g^2)$ is initialized as the running mean and variance $(\mu_s,\sigma_s^2)$ of the pre-trained model.
When adapting the model, we update the global statistics first by exponential moving average as 
\begin{small}
\begin{align}
    \mu_g & = (1-\alpha)\mu_g + \alpha\mu\,, \\
    \sigma_g^2 & = (1-\alpha)\sigma_g^2 + \alpha\sigma^2\,,
\end{align}
\end{small}%
where $(\mu,\sigma^2)$ is the statistics of the buffered samples in the memory bank.  
Then we normalize and affine the feature as Eq.~\eqref{eq:affine} with $(\mu_g,\sigma_g^2)$. When inferring for test samples, we directly utilize $(\mu_g,\sigma_g^2)$ to calculate the output as Eq~\eqref{eq:affine}.
Although simple, RBN is effective enough to tackle the problem of normalization on test streams of \setting.

\paragraph{Category-balanced sampling with timeliness and uncertainty (CSTU).} 
\label{sec:CSTU}
In the \setting setup, the correlation among test samples $\stream_t$ at time $t$ leads to a deviation between the observed distribution $\widehat{\distribution}_{test}$ and the test distribution $\distribution_{test}$. Specifically, the marginal label distribution $p(y|t)$ tends to differ from $p(y)$.
Continuously learning with $\stream_t$ over time $t$ can lead to model adaptation to an unreliable distribution $\widehat{\distribution}_{test}$, resulting in ineffective adaptation and an increased risk of model collapse.

To address this issue, we propose a category-balanced memory bank $\bank$ with a capacity of $\capacity$, which takes into account the timeliness and uncertainty of samples when updating.
In particular, we adopt the predictions of test samples as pseudo labels to guide the update of $\bank$. Meanwhile, to guarantee the balance among categories, we distribute the capacity of $\bank$ equally to each category, and samples of the major categories will be replaced first (refer to lines 5-9 in Algorithm~\ref{alg:sampling}).
Furthermore, due to the continually changing test distribution, old samples in $\bank$ are limited in value, and could even impair the ability of the model to adapt to the current distribution.
Additionally, samples of high uncertainty always produce erroneous gradient information that can hinder model adaptation, as suggested by~\cite{EATA}.

With this in mind, we attach each sample in $\bank$ with a group of heuristics $(\age, \uncertainty)$, where $\age$, initialized as $0$ and increasing with time $t$, is the age of the sample, and $\uncertainty$ the uncertainty calculated as the entropy of the prediction. Next, we combine the timeliness and uncertainty to calculate a heuristic score, i.e., category-balanced sampling with timeliness and uncertainty (CSTU), as follows:
\begin{small}
\begin{equation}
    \label{eq:score}
    \mathcal{H} = \lambda_t \frac{1}{1+\exp(-\age/\capacity)} + \lambda_u \frac{\uncertainty}{\log \numclass}\,,
\end{equation}
\end{small}%
where $\lambda_t$ and $\lambda_u$ make the trade-off between timeliness and uncertainty, and for simplicity, $\lambda_t$ and $\lambda_u$ are set to $1.0$ for all experiments, and $\numclass$ is the number of categories. We summarize our sampling algorithm in Algorithm~\ref{alg:sampling}. With \sampling, we can obtain a robust snapshot of the current test distribution $\distribution_{test}$, and effectively adapt the model to it.

\begin{algorithm}[t]
    \caption{\sampling for one test sample.} 
    \label{alg:sampling}
    \DontPrintSemicolon
    {\bf Input:} a test sample $x$ and the teacher model $f_{\theta^T}$. \\
    {\bf Define:} memory bank $\bank$ and its capacity $\capacity$, number of classes $\numclass$, per class occupation $\mathcal{O}\in \mathbf{R}^\numclass$, total occupation $\varOmega$, classes to pop instance $\mathcal{D}$. \\
    Infer as $p(y|x) = \text{Softmax}(f_{\theta^T}(x))$. \\
    Calculate the predicted category of $x$ as $\hat{y} = \argmax_c{p(c|x)}$, the uncertainty as $\uncertainty_x = -\sum_{c=1}^{C}p(c|x)\log(p(c|x))$, the age as $\age_x=0$, and the heuristic score {\color{blue}$\mathcal{H}_x$} of $x$ with Eq~\eqref{eq:score}\\
    \If{$\mathcal{O}_{\hat{y}} < \frac{\capacity}{\numclass}$}
    {
        if $\varOmega < \capacity$: Search range $\mathcal{D}=\emptyset$. \\
        else: Search range $\mathcal{D}=\{j|j=\argmax_c \mathcal{O}_{c}\}$ \\
    }
    \Else
    {
        Search range $\mathcal{D}=\{\hat{y}\}$ \\
    }
    \If{$\mathcal{D} \text{ is } \emptyset$}
    {
        Add {\color{blue}$(x, \hat{y}, \mathcal{H}_x, \uncertainty_x)$} into $\bank$. \\
    }
    \Else
    {
        Find the instance {\color{red}$(\hat{x} , y_{\hat{x}} ,\age_{\hat{x}} , \uncertainty_{\hat{x}})$} with the highest value in Eq~\eqref{eq:score} {\color{red}$\mathcal{H}_{\hat{x}}$} among $\mathcal{D}$.\\
        \If{{\color{blue}$\mathcal{H}_{x}$} $<$ {\color{red}$\mathcal{H}_{\hat{x}}$}}
        {
            Remove {\color{red}$(\hat{x} , y_{\hat{x}} ,\age_{\hat{x}} , \uncertainty_{\hat{x}})$} from $\bank$.\\
            Add {\color{blue}$(x, \hat{y}, \mathcal{H}_x, \uncertainty_x)$} into $\bank$.
        }
        \Else
        {
            Discard $x$.
        }
    }
    Increase the age of all instances in $\bank$.
\end{algorithm} 

\paragraph{Robust training with timeliness.}
\label{sec:training}
Actually, after replacing BN layers with our RBN and obtaining the memory bank selected via \sampling, we can directly adopt the widely used techniques like pseudo labeling or entropy minimization to perform test-time adaptation. 
However, we notice that too old or unreliable instances still have the opportunity to stay in $\bank$ since keeping the category balance is assigned the top priority. 
In addition, too aggressive updates of the model will make the category balance of $\bank$ unreliable, resulting in unstable adaptation.
Meanwhile, error accumulation caused by the distribution change also makes the aforementioned approaches unworkable.

To further reduce the risk of error gradients information from old and unreliable instances and stabilize the adaptation, we turn to the robust unsupervised learning method teacher-student model and propose a timeliness reweighting strategy. In addition, for the sake of time efficiency and stability, only affine parameters in RBN are trained during adaptation.

At time step $t$, after inferring for the correlated data $\stream_t$ with the teacher model $f_{\theta_t^T}$ and updating the memory bank $\bank$ with $\stream_t$, we begin updating the student model $f_{\theta_t^S}$ and the teacher model $f_{\theta_t^T}$. 
Firstly, we update parameters of student model $\theta_t^S\to\theta_{t+1}^S$ by minimizing the following loss:
\begin{small}
    \begin{equation}
        \label{eq:robust_loss}
        \loss_r=\frac{1}{\varOmega}\sum_{i=1}^{\varOmega}\loss (x_i^{\bank}, \age_i;\theta_t^T,\theta_t^S)\,,
    \end{equation}%
\end{small}%
where $\varOmega=|\bank|$ is the total occupation of the memory bank, and $x_i^{\bank} \text{ and } \age_i (i=1,...,\varOmega)$ are instances in the memory bank and their age respectively. 
Subsequently, the teacher model is updated by exponential moving average as
\begin{small}
    \begin{equation}
        \theta_{t+1}^T=(1-\nu)\theta_{t}^T+\nu\theta_{t+1}^S\,.
    \end{equation}
\end{small}%
To calculate the loss value of an instance $x_i^{\bank}$ from the memory bank, the timeliness reweighting term is computed as
\begin{small}
\begin{equation}
    \label{eq:timeliness}
    E(\age_i)=\frac{\exp(-\age_i/\capacity)}{1+\exp(-\age_i/\capacity)} \,,
\end{equation}
\end{small}%
where $\age_i$ is the age of $x_i^{\bank}$, and $\capacity$ is the capacity of the bank. And then we calculate the cross entropy between the soft-max prediction $p_S(y|x_i'')$ of the strong-augmented view $x_i''$ from the student model and that $p_T(y|x_i')$ of the weak-augmented view~\footnote{Weak augmentation is ReSize+CenterCrop. Strong augmentation is a combination
nine operations like Clip, ColorJitter, and RandomAffine.} $x_i'$ from the teacher model as follows:
\begin{small}
\begin{equation}
    \label{eq:ce}
    \ell(x_i',x_i'') = -\frac{1}{\numclass}\sum_{c=1}^{\numclass}p_T(c|x_i')\log p_S(c|x_i'') \,.
\end{equation}%
\end{small}
Finally, equipped with Eq.~\eqref{eq:timeliness} and Eq.~\eqref{eq:ce}, the right-hand side of Eq.~\eqref{eq:robust_loss} reduces to
\begin{small}
\begin{equation}
    \loss (x_i^{\bank}, \age_i;\theta_t^T,\theta_t^S)=E(\age_i)\ell(x_i',x_i'')\,.
\end{equation}%
\end{small}%
To sum up, equipped with RBN, CSTU, and robust training with timeliness, our \method is capable of effectively adapting any pre-trained models in dynamic scenarios.

\section{Experiments}
\label{sec:Experiment}
\subsection{Setup}
\paragraph{Datasets.} CIFAR10-C and CIFAR100-C~\cite{corruptions} are the commonly used TTA benchmarks to testify the robustness under corruptions. Both of them are obtained by applying 15 kinds of corruption with 5 different degrees of severity on their clean test images of original datasets CIFAR10 and CIFAR100 respectively. CIFAR10/CIFAR100~\cite{cifar} have 50,000/10,000 training/test images, all of which fall into 10/100 categories.
DomainNet~\cite{DomainNet} is the largest and hardest dataset to date for domain adaptation and consists of about 0.6 million images with 345 classes. It consists of six different domains including {Clipart} ({clp}), {Infograph} ({inf}), {Painting} ({pnt}), {Quickdraw} ({qdr}), {Real} ({rel}), and {Sketch} ({skt}). 
We first pre-train a source model on the train set in one of six domains and testify all baseline methods on the test set of the remaining five domains.

\begin{table*}[t]
    \centering
    \caption{Average classification error of the task CIFAR10 $\to$ CIFAR10-C while continually adapting to different corruptions at the highest severity 5 with correlatively sampled test stream under the proposed setup \setting.
    }
    \label{table:cifar10}
    \VspaceBefore
    \resizebox{\textwidth}{!}{
    \renewcommand{\arraystretch}{0.8}
    {
    \begin{tabular}{l|ccccccccccccccc|c}
        \toprule[1.2pt]
        Time & \multicolumn{15}{l|}{$t\xrightarrow{\hspace*{18.5cm}}$}& \\ \hline
        Method & \rotatebox[origin=c]{45}{motion} & \rotatebox[origin=c]{45}{snow} & \rotatebox[origin=c]{45}{fog} & \rotatebox[origin=c]{45}{shot} & \rotatebox[origin=c]{45}{defocus} & \rotatebox[origin=c]{45}{contrast} & \rotatebox[origin=c]{45}{zoom} & \rotatebox[origin=c]{45}{brightness} & \rotatebox[origin=c]{45}{frost} & \rotatebox[origin=c]{45}{elastic} & \rotatebox[origin=c]{45}{glass} & \rotatebox[origin=c]{45}{gaussian} & \rotatebox[origin=c]{45}{pixelate} & \rotatebox[origin=c]{45}{jpeg} & \rotatebox[origin=c]{45}{impulse}
        & Avg. \\ 
        
        \midrule
        Source & 34.8 & 25.1 & 26.0 & 65.7 & 46.9 & 46.7 & 42.0 & \underline{9.3} & 41.3 & 26.6 & 54.3 & 72.3 & 58.5 & 30.3 & 72.9 & 43.5\\

        BN~\cite{BN_Stat} & 73.2 & 73.4 & 72.7 & 77.2 & 73.7 & 72.5 & 72.9 & 71.0 & 74.1 & 77.7 & 80.0 & 76.9 & 75.5 & 78.3 & 79.0 & 75.2 \\

        PL~\cite{PL} & 73.9 & 75.0 & 75.6 & 81.0 & 79.9 & 80.6 & 82.0 & 83.2 & 85.3 & 87.3 & 88.3 & 87.5 & 87.5 & 87.5 & 88.2 & 82.9 \\

        TENT~\cite{tent_wang2020} & 74.3 & 77.4 & 80.1 & 86.2 & 86.7 & 87.3 & 87.9 & 87.4 & 88.2 & 89.0 & 89.2 & 89.0 & 88.3 & 89.7 & 89.2 & 86.0 \\
        
        LAME~\cite{niid_boudiaf2022parameter} & 29.5 & \bf 19.0 & \underline{20.3} & 65.3 & 42.4 & 43.4 & 36.8 & \bf 5.4 & 37.2 & \bf 18.6 & 51.2 & 73.2 & 57.0 & \bf 22.6 & 71.3 & 39.5 \\

        CoTTA~\cite{cotta} & 77.1 & 80.6 & 83.1 & 84.4 & 83.9 & 84.2 & 83.1 & 82.6 & 84.4 & 84.2 & 84.5 & 84.6 & 82.7 & 83.8 & 84.9 & 83.2 \\

        NOTE~\cite{note} & \bf 18.0 & 22.1 & 20.6 & \underline{35.6} & \underline{26.9} & \bf 13.6 & \underline{26.5} & 17.3 & \underline{27.2} & 37.0 & \underline{48.3} & \underline{38.8} & \underline{42.6} & 41.9 & \underline{49.7} & \underline{31.1} \\

        \midrule

        \method & \underline{18.1} & \underline{21.3} & \bf 18.8 & \bf 33.6 & \bf 23.6 & \underline{16.5} & \bf 15.1 & 11.2 & \bf 21.9 & \underline{30.7} & \bf 39.6 & \bf 26.8 & \bf 33.7 & \underline{27.8} & \bf 39.5 & \bf 25.2\dtplus{+5.9} \\

    \bottomrule[1.2pt]
    \end{tabular}
    }
    }
    \VspaceAfter
\end{table*}

\begin{table*}[t]
    \centering
    \caption{Average classification error of the task CIFAR100 $\to$ CIFAR100-C while continually adapting to different corruptions at the highest severity 5 with correlatively sampled test stream under the proposed setup \setting.
    }
    \label{table:cifar100}
    \VspaceBefore
    \resizebox{\linewidth}{!}{
    \renewcommand{\arraystretch}{0.8}
    {
    \begin{tabular}{l|ccccccccccccccc|c}
        \toprule[1.2pt]
        Time & \multicolumn{15}{l|}{$t\xrightarrow{\hspace*{18.5cm}}$}& \\ \hline
        Method & \rotatebox[origin=c]{45}{motion} & \rotatebox[origin=c]{45}{snow} & \rotatebox[origin=c]{45}{fog} & \rotatebox[origin=c]{45}{shot} & \rotatebox[origin=c]{45}{defocus} & \rotatebox[origin=c]{45}{contrast} & \rotatebox[origin=c]{45}{zoom} & \rotatebox[origin=c]{45}{brightness} & \rotatebox[origin=c]{45}{frost} & \rotatebox[origin=c]{45}{elastic} & \rotatebox[origin=c]{45}{glass} & \rotatebox[origin=c]{45}{gaussian} & \rotatebox[origin=c]{45}{pixelate} & \rotatebox[origin=c]{45}{jpeg} & \rotatebox[origin=c]{45}{impulse}
        & Avg. \\ 
        
        \midrule
        Source & 30.8 & 39.5 & 50.3 & 68.0 & \underline{29.3} & 55.1 & 28.8 & 29.5 & 45.8 & 37.2 & 54.1 & 73.0 & 74.7 & 41.2 & \underline{39.4} & 46.4 \\

        BN~\cite{BN_Stat} & 48.5 & 54.0 & 58.9 & 56.2 & 46.4 & \underline{48.0} & 47.0 & 45.4 & 52.9 & 53.4 & 57.1 & 58.2 & 51.7 & 57.1 & 58.8 & 52.9 \\

        PL~\cite{PL} & 50.6 & 62.1 & 73.9 & 87.8 & 90.8 & 96.0 & 94.8 & 96.4 & 97.4 & 97.2 & 97.4 & 97.4 & 97.3 & 97.4 & 97.4 & 88.9 \\

        TENT~\cite{tent_wang2020} & 53.3 & 77.6 & 93.0 & 96.5 & 96.7 & 97.5 & 97.1 & 97.5 & 97.3 & 97.2 & 97.1 & 97.7 & 97.6 & 98.0 & 98.3 & 92.8 \\

        LAME~\cite{niid_boudiaf2022parameter} & \bf 22.4 & \bf 30.4 & \underline{43.9} & 66.3 & \bf 21.3 & 51.7 & \bf 20.6 & \bf 21.8 & \underline{39.6} & \bf 28.0 & \underline{48.7} & 72.8 & 74.6 & \bf 33.1 & \bf 32.3 & \underline{40.5} \\

        CoTTA~\cite{cotta} & 49.2 & 52.7 & 56.8 & \underline{53.0} & 48.7 & 51.7 & 49.4 & 48.7 & 52.5 & 52.2 & 54.3 & \underline{54.9} & \underline{49.6} & 53.4 & 56.2 & 52.2 \\

        NOTE~\cite{note} & 45.7 & 53.0 & 58.2 & 65.6 & 54.2 & 52.0 & 59.8 & 63.5 & 74.8 & 91.8 & 98.1 & 98.3 & 96.8 & 97.0 & 98.2 & 73.8  \\
    
        \midrule

        \method & \underline{31.8} & \underline{36.7} & \bf 40.9 & \bf 42.1 & 30.0 & \bf 33.6 & \underline{27.9} & \underline{25.4} & \bf 32.3 & \underline{34.0} & \bf 38.8 & \bf 38.7 & \bf 31.3 & \underline{38.0} & 42.9 & \bf 35.0\dtplus{+5.5} \\
    
    \bottomrule[1.2pt]
    \end{tabular}
    }
    }
    \VspaceAfter
\end{table*}

\paragraph{Implementation details.}
All experiments are conducted with PyTorch~\cite{paszke2019pytorch} framework. In the case of robustness to corruption, following the previous methods~\cite{tent_wang2020, EATA, cotta}, we obtain the pre-trained model from RobustBench benchmark~\cite{RobustBench}, including the WildResNet-28~\cite{wildresnet} for CIFAR10 $\to$ CIFAR10-C, and the ResNeXt-29~\cite{ResNext} for CIFAR100 $\to$ CIFAR100-C.
Then, we change the test corruption at the highest severity 5 one by one to simulate that the test distribution continually changes with time in \setting. And in the case of generalization under the huge domain gap, we train a ResNet-101~\cite{resnet} by standard classification loss for each domain in DomainNet and adapt them continually to different domains except the source domain. Meanwhile, we utilize the Dirichlet distribution to simulate the correlatively sampled test stream for all datasets.
For optimization, we adopt Adam~\cite{Adam} optimizer with learning rate $1.0\times 10^{-3}$, $\beta=0.9$. For a fair comparison, we set the batch size for all methods as $64$ and the capacity of the memory bank of \method as $\capacity =64$. Concerning the hyperparameters, we adopt a unified set of values for \method across all experiments including $\alpha=0.05$, $\nu=0.001$, $\lambda_t=1.0$, $\lambda_u=1.0$, and $\delta=0.1$. More details are provided in the appendix.

\begin{table*}[!htbp]
    \caption{Average classification error of DomainNet while continually adapting to different domains with correlatively sampled test stream.}
    \VspaceBefore
    \label{table:domainnet}
   \setlength{\tabcolsep}{0.085em}
   \resizebox{\linewidth}{!}{
   \renewcommand{\arraystretch}{0.8}
   {
   \centering
   
   \begin{tabular}{c|c c c c c c c || c | c c c c c c c || c |c c c c c c c || c | c c c c c c c }
    \toprule[1.2pt]

    Time & \multicolumn{6}{c}{$t\xrightarrow{\hspace*{3.5cm}}$} & & Time & \multicolumn{6}{c}{$t\xrightarrow{\hspace*{3.5cm}}$} & & Time & \multicolumn{6}{c}{$t\xrightarrow{\hspace*{3.5cm}}$} & & Time & \multicolumn{6}{c}{$t\xrightarrow{\hspace*{3.5cm}}$} &  \\

    \hline

   \redbold{Source} & \BV{clp} & \BV{inf} & \BV{pnt} & \BV{qdr} & \BV{rel} & \BV{skt} & \BU{Avg.} &
   \redbold{BN} & \BV{clp} & \BV{inf} & \BV{pnt} & \BV{qdr} & \BV{rel} & \BV{skt} & \BU{Avg.} &
   \redbold{PL} & \BV{clp} & \BV{inf} & \BV{pnt} & \BV{qdr} & \BV{rel} & \BV{skt} & \BU{Avg.} &
   \redbold{TENT} & \BV{clp} & \BV{inf} & \BV{pnt} & \BV{qdr} & \BV{rel} & \BV{skt} & \BU{Avg.} \\
   \midrule
   
   \BV{clp}& N/A& 83.9 & 65.4 & 88.6 & 48.0 & 59.1 & 69.0 & 
   \BV{clp}& N/A& 88.6 & 70.7 & 90.5 & 65.4 & 67.0 & 76.5 & 
   \BV{clp}& N/A& 94.5 & 98.9 & 99.5 & 99.7 & 99.7 & 98.5 & 
   \BV{clp}& N/A& 87.5 & 71.9 & 94.2 & 96.2 & 98.9 & 89.7 \\

    \BV{inf}& 61.8 & N/A& 66.9 & 96.0 & 50.0 & 70.6 & 69.1 & 
    \BV{inf}& 68.6 & N/A& 74.2 & 96.2 & 69.9 & 76.8 & 77.1 & 
    \BV{inf}& 82.6 & N/A& 99.2 & 99.6 & 99.7 & 99.3 & 96.1 & 
    \BV{inf}& 68.6 & N/A& 75.0 & 97.3 & 95.9 & 98.7 & 87.1 \\

    \BV{pnt}& 56.5 & 83.7 & N/A& 94.2 & 42.6 & 63.4 & 68.1 & 
    \BV{pnt}& 60.8 & 87.9 & N/A& 94.3 & 62.3 & 68.7 & 74.8 & 
    \BV{pnt}& 78.6 & 99.4 & N/A& 99.7 & 99.6 & 99.7 & 95.4 & 
    \BV{pnt}& 61.7 & 87.1 & N/A& 96.4 & 95.3 & 98.8 & 87.8 \\

    \BV{qdr}& 89.2 & 99.0 & 98.6 & N/A& 95.0 & 92.3 & 94.8 & 
    \BV{qdr}& 80.3 & 97.7 & 92.6 & N/A& 88.7 & 88.1 & 89.5 & 
    \BV{qdr}& 81.7 & 99.5 & 99.6 & N/A& 99.7 & 99.8 & 96.1 & 
    \BV{qdr}& 78.9 & 97.1 & 91.6 & N/A& 89.2 & 88.7 & 89.1 \\

    \BV{rel}& 49.4 & 80.4 & 51.5 & 93.4 & N/A& 63.3 & 67.6 & 
    \BV{rel}& 57.9 & 87.1 & 63.1 & 94.3 & N/A& 70.8 & 74.6 & 
    \BV{rel}& 73.5 & 99.4 & 99.2 & 99.6 & N/A& 99.7 & 94.3 & 
    \BV{rel}& 57.8 & 86.4 & 68.1 & 96.9 & N/A& 96.7 & 81.2 \\

    \BV{skt}& 47.5 & 88.2 & 62.9 & 87.1 & 51.8 & N/A& 67.5 & 
    \BV{skt}& 50.4 & 87.6 & 64.6 & 89.6 & 63.1 & N/A& 71.1 & 
    \BV{skt}& 64.8 & 99.2 & 99.4 & 99.7 & 99.7 & N/A& 92.6 & 
    \BV{skt}& 51.9 & 87.2 & 69.1 & 95.3 & 97.3 & N/A& 80.1 \\

    \BU{Avg.}& 60.9 & 87.0 & 69.1 & 91.9 & 57.5 & 69.7 & 72.7 & 
    \BU{Avg.}& 63.6 & 89.8 & 73.0 & 93.0 & 69.9 & 74.3 & 77.3 & 
    \BU{Avg.}& 76.2 & 98.4 & 99.3 & 99.6 & 99.7 & 99.6 & 95.5 & 
    \BU{Avg.}& 63.8 & 89.0 & 75.1 & 96.0 & 94.8 & 96.4 & 85.8 \\

    \hline\hline

   Time & \multicolumn{6}{c}{$t\xrightarrow{\hspace*{3.5cm}}$} & & Time & \multicolumn{6}{c}{$t\xrightarrow{\hspace*{3.5cm}}$} & & Time & \multicolumn{6}{c}{$t\xrightarrow{\hspace*{3.5cm}}$} & & Time & \multicolumn{6}{c}{$t\xrightarrow{\hspace*{3.5cm}}$} & \\

    \hline
   
   \redbold{LAME} & \BV{clp} & \BV{inf} & \BV{pnt} & \BV{qdr} & \BV{rel} & \BV{skt} & \BU{Avg.} &
   \redbold{COTTA} & \BV{clp} & \BV{inf} & \BV{pnt} & \BV{qdr} & \BV{rel} & \BV{skt} & \BU{Avg.} &
   \redbold{NOTE} & \BV{clp} & \BV{inf} & \BV{pnt} & \BV{qdr} & \BV{rel} & \BV{skt} & \BU{Avg.} &
   \redbold{\method} & \BV{clp} & \BV{inf} & \BV{pnt} & \BV{qdr} & \BV{rel} & \BV{skt} & \BU{Avg.} \\
   
   \midrule
   
   \BV{clp}& N/A& 82.2 & 64.5 & 87.7 & 46.9 & 58.9 & 68.0 & 
   \BV{clp}& N/A& 90.6 & 77.9 & 89.3 & 76.3 & 72.7 & 81.4 & 
   \BV{clp}& N/A& 89.2 & 73.0 & 94.8 & 98.4 & 99.4 & 91.0 & 
   \BV{clp}& N/A& 85.5 & 62.0 & 82.0 & 49.3 & 59.8 & 67.7 \\

    \BV{inf}& 60.1 & N/A& 65.7 & 95.4 & 48.5 & 69.4 & 67.8 & 
    \BV{inf}& 74.5 & N/A& 82.0 & 95.7 & 80.2 & 81.5 & 82.8 & 
    \BV{inf}& 75.4 & N/A& 78.7 & 98.7 & 98.1 & 99.5 & 90.1 & 
    \BV{inf}& 61.8 & N/A& 63.7 & 91.5 & 52.5 & 67.6 & 67.4 \\

    \BV{pnt}& 55.8 & 81.5 & N/A& 93.3 & 41.3 & 62.1 & 66.8 & 
    \BV{pnt}& 66.3 & 89.8 & N/A& 93.4 & 74.0 & 75.4 & 79.8 & 
    \BV{pnt}& 64.7 & 89.8 & N/A& 97.8 & 98.4 & 99.2 & 90.0 & 
    \BV{pnt}& 53.3 & 84.1 & N/A& 89.1 & 47.3 & 61.4 & 67.0 \\

    \BV{qdr}& 88.3 & 99.1 & 99.0 & N/A& 94.9 & 92.2 & 94.7 & 
    \BV{qdr}& 82.3 & 98.2 & 94.6 & N/A& 92.5 & 90.1 & 91.5 & 
    \BV{qdr}& 74.7 & 97.2 & 92.2 & N/A& 93.5 & 99.6 & 91.4 & 
    \BV{qdr}& 77.5 & 97.0 & 89.8 & N/A& 80.3 & 82.2 & 85.3 \\

    \BV{rel}& 48.0 & 79.3 & 50.1 & 91.6 & N/A& 60.2 & 65.8 & 
    \BV{rel}& 64.0 & 90.3 & 73.2 & 93.5 & N/A& 77.6 & 79.7 & 
    \BV{rel}& 61.3 & 89.2 & 68.9 & 98.8 & N/A& 99.2 & 83.5 & 
    \BV{rel}& 49.1 & 82.3 & 50.3 & 88.0 & N/A& 61.1 & 66.2 \\

    \BV{skt}& 45.6 & 87.1 & 59.5 & 83.9 & 49.9 & N/A& 65.2 & 
    \BV{skt}& 56.1 & 89.2 & 71.9 & 89.2 & 73.5 & N/A& 76.0 & 
    \BV{skt}& 55.2 & 89.7 & 70.1 & 96.9 & 98.3 & N/A& 82.0 & 
    \BV{skt}& 42.6 & 83.7 & 54.4 & 80.9 & 47.5 & N/A& 61.8 \\

    \BU{Avg.}& 59.6 & 85.8 & 67.8 & 90.4 & 56.3 & 68.6 & \underline{71.4} & 
    \BU{Avg.}& 68.6 & 91.6 & 79.9 & 92.2 & 79.3 & 79.5 & 81.9 & 
    \BU{Avg.}& 66.3 & 91.0 & 76.6 & 97.4 & 97.3 & 99.4 & 88.0 & 
    \BU{Avg.}& 56.8 & 86.5 & 64.0 & 86.3 & 55.4 & 66.4 & \bf 69.2\dtplus{+2.2} \\

   \bottomrule[1.2pt]
   
   \end{tabular}
   }
   }
   \VspaceAfter
\end{table*}

\begin{figure*}[t]
    \centering
    \subfloat[CIFAR10-C.]{
      \begin{minipage}[b]{0.24\linewidth} 
        \centering  
        \includegraphics[width=\linewidth]{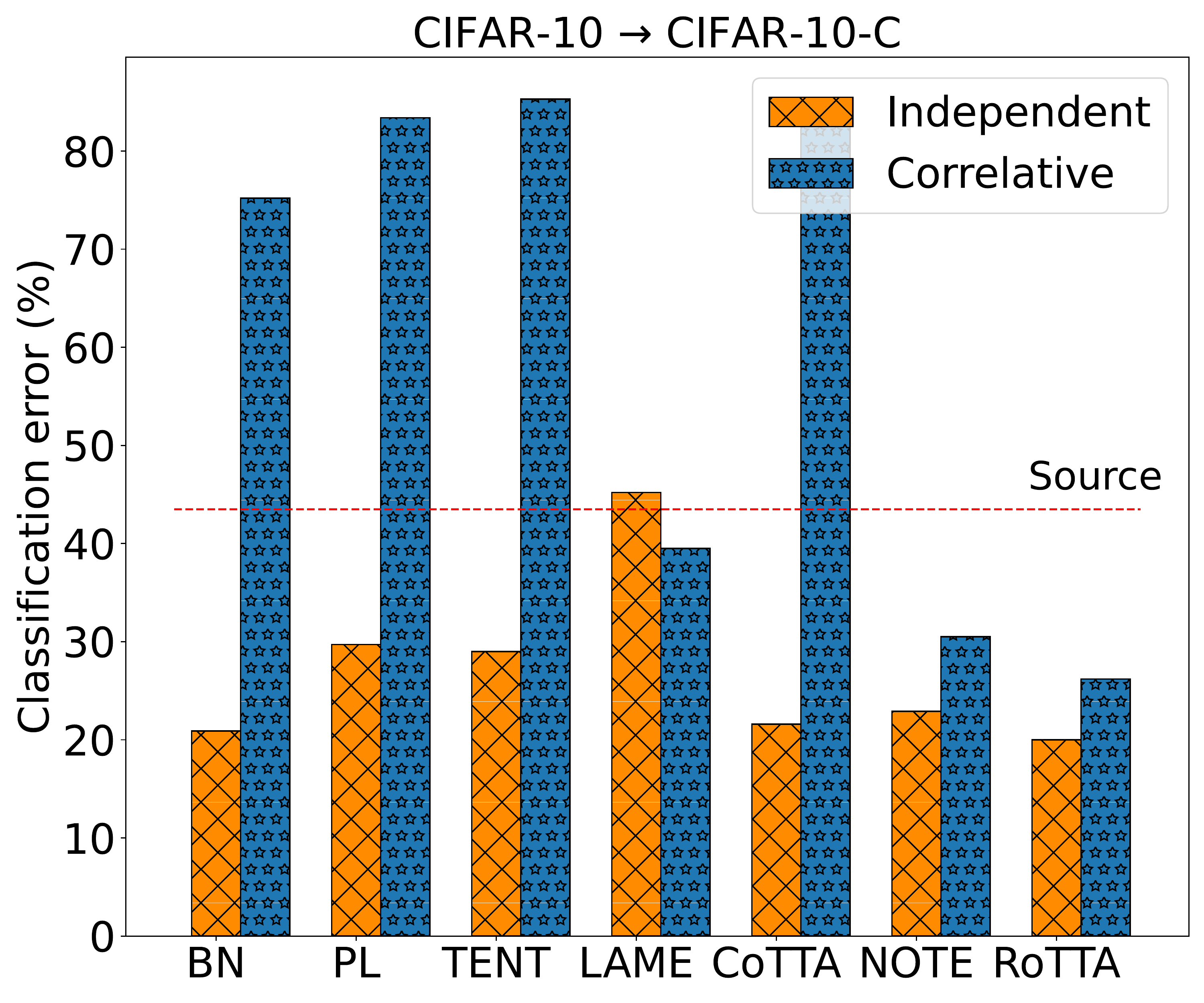} 
        \vspace{-4mm}
        \label{fig:cifar10_10ordef}
      \end{minipage}
    }
    \subfloat[CIFAR100-C.]{
      \begin{minipage}[b]{0.24\linewidth} 
        \centering  
        \includegraphics[width=\linewidth]{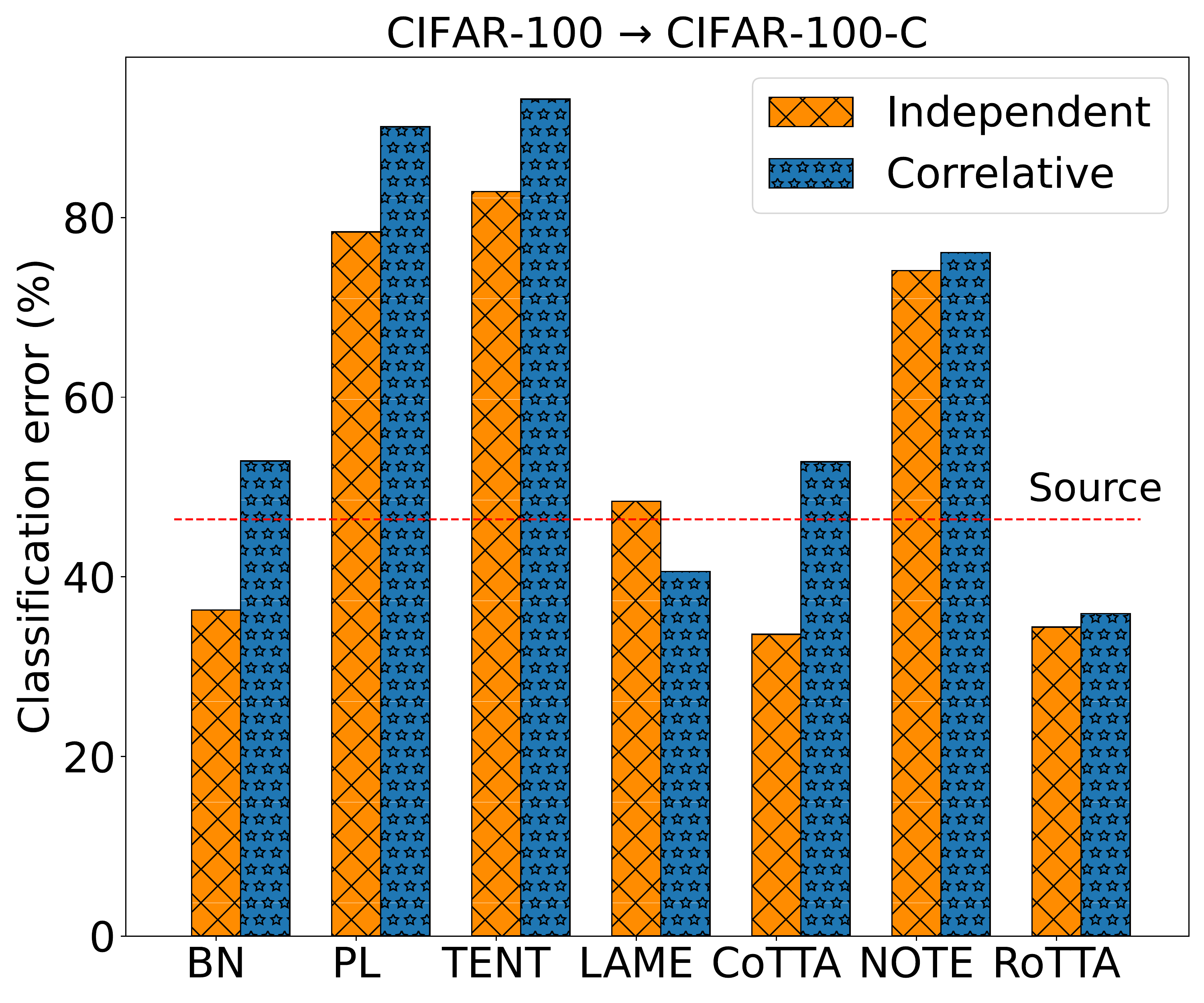} 
        \vspace{-4mm}
        \label{fig:cifar100_10ordef}
      \end{minipage}%
    }
    \subfloat[$\delta$.]{
      \begin{minipage}[b]{0.24\linewidth} 
        \centering  
        \includegraphics[width=\linewidth]{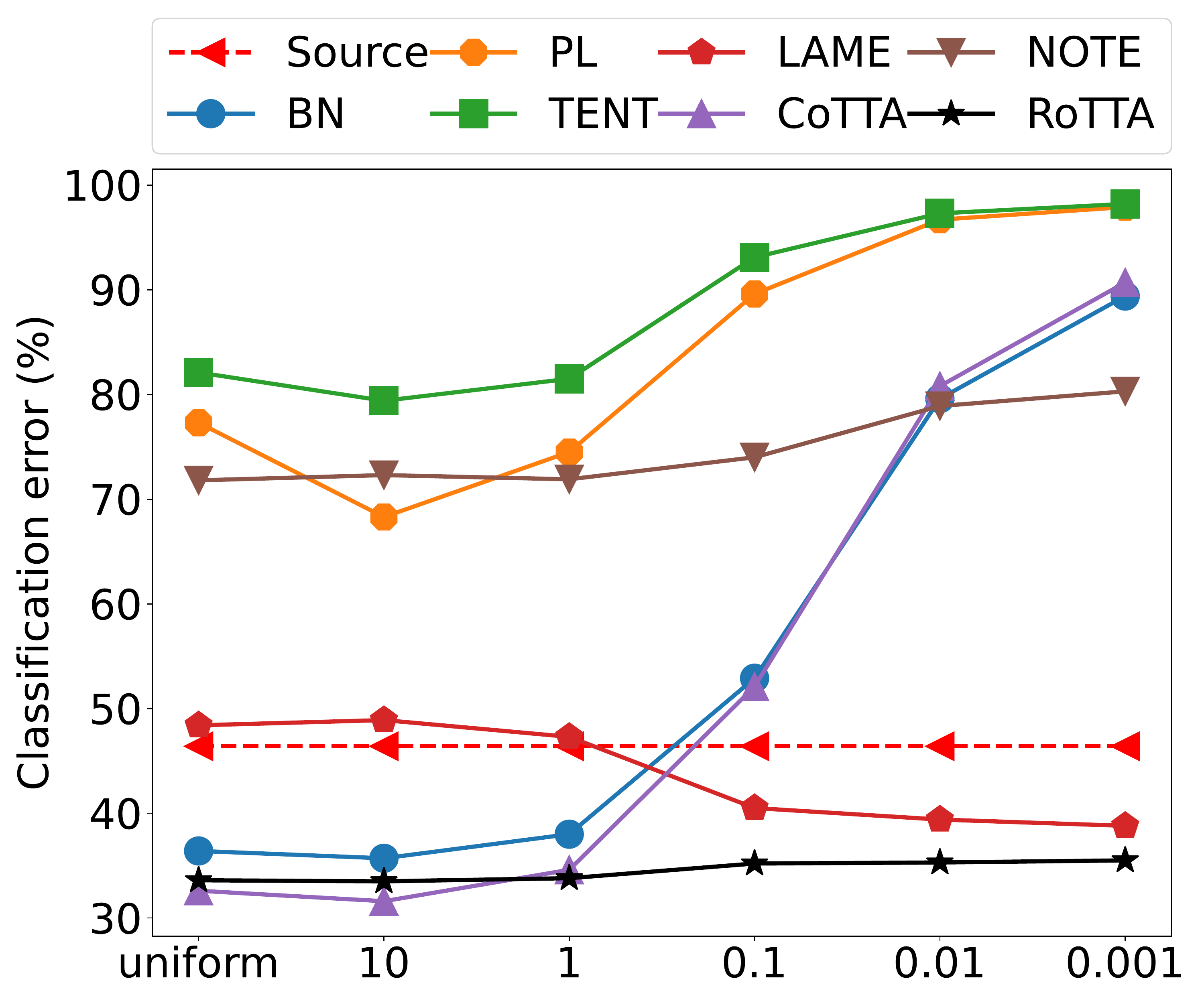}   \vspace{-4mm}
        \label{fig:analyze_delta}
      \end{minipage}
    }
    \subfloat[Batch size.]{
      \begin{minipage}[b]{0.24\linewidth} 
        \centering  
        \includegraphics[width=\linewidth]{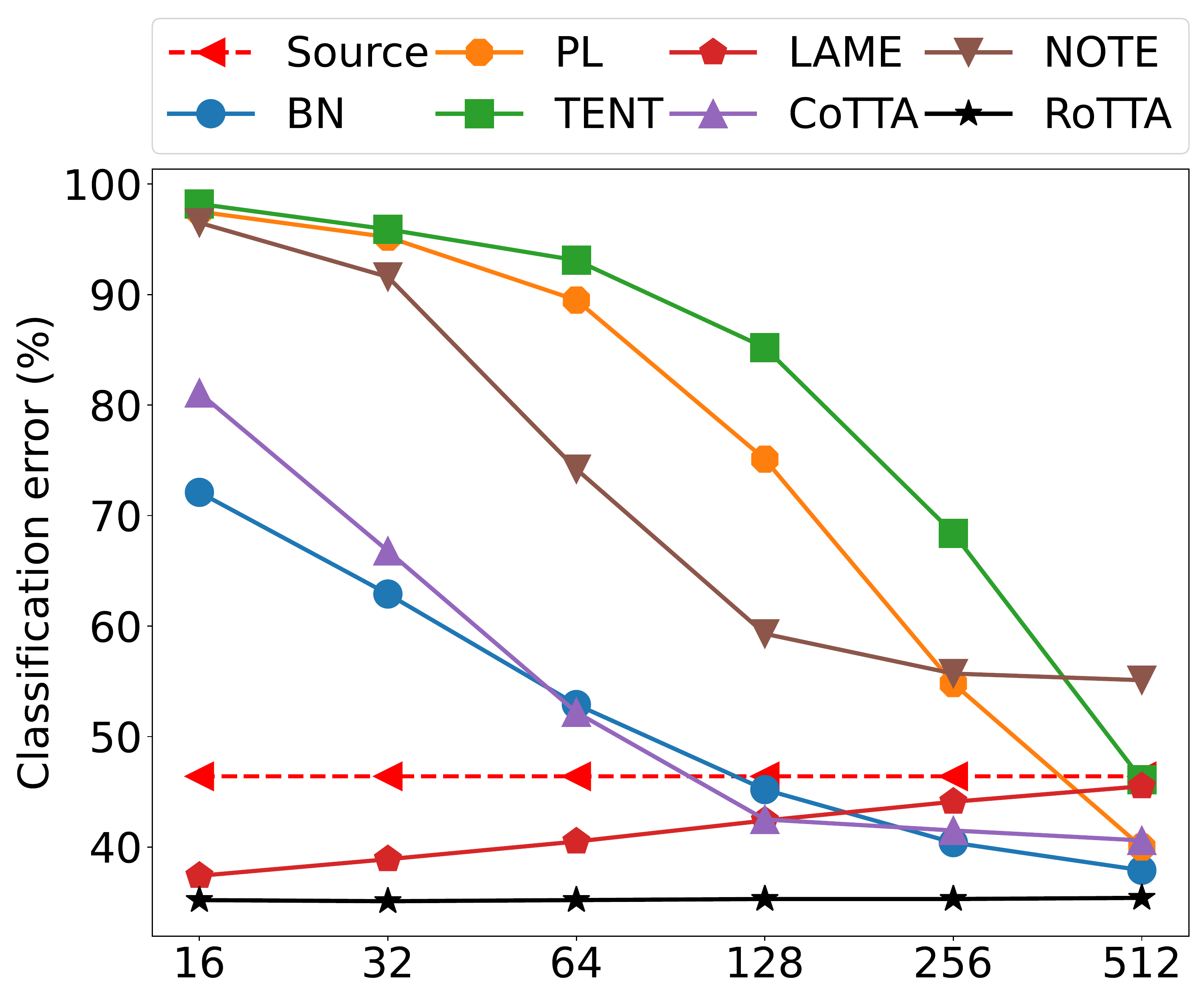}    \vspace{-4mm}
        \label{fig:analyze_bs}
      \end{minipage}%
    }
    \VspaceBefore
    \caption{(a) \& (b) we adapt the model continually to different corruptions of 10 different orders with independently and correlatively sampled test streams on CIFAR10-C and CFAR100-C respectively and report their average classification error. (c) \& (d) we verify the effect of $\delta$ and batch size to different methods on CIFAR100-C respectively. }
    \label{fig_cifar_params}
    \vspace{-4mm}
\end{figure*} 

\subsection{Comparisons with the State-of-the-arts}
\paragraph{Robustness under corruptions.} The classification error on CIFAR10$\to$CIFAR10-C and CIFAR100$\to$CIFAR100-C are shown in Table~\ref{table:cifar10} and Table~\ref{table:cifar100} respectively. We change the type of the current corruption at the highest severity 5 as time goes on, and sample data correlatively for inference and adaptation simultaneously. The same test stream is shared across all compared methods. 

From Table~\ref{table:cifar10} and Table~\ref{table:cifar100},  we can see that \method achieves the best performance compared to previous methods. Moreover, \method has a significant performance gain to the second-best method that \gain{$\bf 5.9\%$} improvement on CIFAR10$\to$CIFAR10-C and \gain{$\bf 5.5\%$} improvement on CIFAR100$\to$CIFAR100-C respectively, verifying the effectiveness of \method to adapt the model under \setting.

In more detail, we can observe that BN~\cite{BN_Stat}, PL~\cite{PL}, TENT~\cite{tent_wang2020} and CoTTA~\cite{cotta} negatively adapt the model to the test streams of both datasets compared to Source ($-6.5\sim-46.4\%$). 
This is attributed to the fact that these methods overlook the issues posed by correlation sampling, which can result in highly correlated data within a batch. As a consequence, traditional normalization statistics may be ineffective in appropriately normalizing the feature maps.
Equipped with RBN and \sampling, \method no longer suffers from this issue. Meanwhile, in Table~\ref{table:cifar100}, if focus on the adaptation procedure, we can see that the performance of PL~\cite{PL}, TENT~\cite{tent_wang2020} and NOTE~\cite{note} becomes worse and worse, and eventually, the model even collapses (error rate $>$ 97\%). This reveals that the impact of error accumulation on long-term adaptation can be catastrophic. To tackle this problem, \method turns to robustly adapt the model with timeliness reweighting and confident samples in the memory bank, and superior performance throughout the adaptation process demonstrates its effectiveness. 

In addition, we find that although LAME~\cite{niid_boudiaf2022parameter} never tunes the parameters of the model, it is still a competitive baseline for example it achieves the second-best result on CIFAR100$\to$CIFAR100-C. However, its performance is very dependent on the performance of the pre-trained model e.g. negligible improvement on difficult corruptions (shot, gaussian, pixelate). On the contrary, our \method is more flexible and achieves better and more robust results.

\paragraph{Generalization under domain shift.} We also evaluate \method under a more challenging dataset DomainNet, where we continually adapt a source pre-trained model to correlatively sampled test streams of the rest domains. As shown in Table~\ref{table:domainnet}, consistent with the previous analysis, most of the methods include BN~\cite{BN_Stat}, PL~\cite{PL}, TENT~\cite{tent_wang2020}, CoTTA~\cite{cotta} and NOTE~\cite{note} even perform worse than the Source model ($-4.6\sim-22.8\%$). \method consistently achieves the best performance and has \gain{$\bf 2.2\%$} gain than the second method LAME~\cite{niid_boudiaf2022parameter}, demonstrating \method's effectiveness again.

\subsection{Ablation Study}
\label{sec:ablation}
\paragraph{Effect of each component.} To further investigate the efficacy of each component, we replace each part with the normally used solutions to obtain three variants: (1) \method w/o RBN, replace RBN with test-time BN in TENT~\cite{tent_wang2020}; (2) \method w/o CSTU, directly adapt the model on test stream; (3) \method w/o robust training (RT), directly adapt the model only with entropy minimization. As shown in Table~\ref{table:ablation}, we can observe that significant performance degradation occurs for all variants, proving that every part of our proposed method is valid for \setting. Take one component for a detailed example, without RBN robustly normalizing feature maps, the performance of \method drops $50.2\%$ and $16.3\%$ on CIFAR10-C and CIFAR100-C respectively, proving that RBN is robust enough to tackle the problem of normalization of correlatively sampled data streams. CSTU enables \method to adapt to a more stable distribution by maintaining a timely and confident snapshot of the test distribution. Meanwhile, robust training with timeliness greatly reduces the accumulation of errors. Every component behaves significantly to enable effective adaptation under \setting.

\paragraph{Effect of the distribution changing order.} To exclude the effect of a fixed order of distribution changing, we conducted experiments on ten different sequences of changes on CIFAR10-C and CIFAR100-C with independently and correlatively sampled test streams respectively. As shown in Figure~\ref{fig:cifar10_10ordef} and~\ref{fig:cifar100_10ordef}, no matter what kind of setup, \method can achieve excellent results. The detailed results on the correlatively sampled test streams are shown in Table~\ref{table:different_orders}, \method achieves \gain{$\bf 4.3\%$} and \gain{$\bf 4.7\%$} progress on CIFAR10-C and CIFAR100-C respectively. This shows that \method can adapt the model robustly and effectively in long-term scenarios where distribution continually changes and test streams are sampled either independently or correlatively, making it a good choice for model deployment.

\paragraph{Effect of Dirichlet concentration parameter $\delta$.} We vary the value of $\delta$ on CIFAR100-C and compare \method with other approaches in Figure~\ref{fig:analyze_delta}. As the value of $\delta$ increases, the performance of BN~\cite{BN_Stat}, PL~\cite{PL}, TENT~\cite{tent_wang2020} and CoTTA~\cite{cotta} drops quickly, because they never consider the increasing correlation among test samples. NOTE~\cite{note} is stable to correlatively sampled test streams but does not consider the distribution changing, causing ineffective adaptation. Meanwhile, the higher correlation between test samples will make the propagation of labels more accurate, which is why the result of LAME~\cite{niid_boudiaf2022parameter} slightly improves. Finally, excellent and stable results once again prove the stability and effectiveness of \method.

\begin{table}[t]
    \centering
    \caption{Classification error of different variants of our \method.}
    \label{table:ablation}
    \VspaceBefore
    \resizebox{0.94\linewidth}{!}{
    \renewcommand{\arraystretch}{0.8}
    {
    \begin{tabular}{l|cc|c}
        \toprule[1.2pt]
        Variant & CIFAR10-C & CIFAR100-C & Avg. \\
        \midrule
        \method w/o RBN & 75.4 & 51.3 & 63.4 \\
        \method w/o CSTU & 47.1 & 46.3 & 46.7 \\
        \method w/o RT & 78.2 & 95.0 & 81.6 \\
        \midrule
        \method & \bf 25.2 & \bf 35.0 & \bf 30.1 \\
    \bottomrule[1.2pt]
    \end{tabular}
    }
    }
    \VspaceAfter
\end{table}

\paragraph{Effect of batch size.} In real scenarios, considering deployment environments may use different test batch sizes, we conduct experiments with different values of test batch sizes and results are shown in Figure~\ref{fig:analyze_bs}. For a fair comparison, we control the frequency of updating the model of \method so that the number of samples involved in back-propagation is the same. As the batch size increases, we can see that all of the compared methods have a significant improvement except for lame which has a slight decrease. This is because the number of categories in a batch increases with the increasing batch size, causing the overall correlation to become lower but the propagation of labels to become more difficult. Most significantly, \method achieves the best results across different batch sizes, demonstrating its robustness in dynamic scenarios once again.

\begin{table}[t]
    \centering
    \caption{Average classification error of tasks CIFAR10 $\to$ CIFAR10-C and CIFAR100 $\to$ CIFAR100-C while continually adapting to different corruptions of 10 different orders at the highest severity 5 with correlatively sampled test stream.
    }
    \label{table:different_orders}
    \VspaceBefore
    \resizebox{0.9\linewidth}{!}{
    \renewcommand{\arraystretch}{0.8}
    {
    \begin{tabular}{l|cc|c}
        \toprule[1.2pt]
        Method & CIFAR10-C & CIFAR100-C & Avg. \\
        \midrule
        Source & 43.5 & 46.4 & 46.9 \\
        
        BN~\cite{BN_Stat} & 75.2 & 52.9 & 64.1 \\

        PL~\cite{PL} & 75.2 & 52.9 & 60.1 \\

        TENT~\cite{tent_wang2020} & 82.3 & 93.2 & 87.8 \\

        LAME~\cite{niid_boudiaf2022parameter} & 39.5 & \underline{40.6} & \underline{40.1} \\

        NOTE~\cite{note} & \underline{30.5} & 76.1 & 53.3 \\

        CoTTA~\cite{cotta} & 83.1 & 52.8 & 67.9 \\
    
        \midrule

        \method & \bf 26.2\dtplus{+4.3} & \bf 35.9\dtplus{+4.7} & \bf 31.1\dtplus{+9.0} \\
    
    \bottomrule[1.2pt]
    \end{tabular}
    }
    }
    \VspaceAfter
    \vspace{-2.3mm}
\end{table}

\section{Conclusion}
\label{sec:conclusion}
This work proposes a more realistic TTA setting where distribution changing and correlative sampling occur simultaneously at the test phase, namely Practical Test-Time Adaptation (\setting).  To tackle the problems of \setting, we propose {\bf Ro}bust {\bf T}est-{\bf T}ime {\bf A}daptation (\method) method against the complex data stream. More specifically, a group of robust statistics for the normalization of feature maps is estimated by robust batch normalization. Meanwhile, a memory bank is adopted to capture a snapshot of the test distribution by category-balanced sampling with considering timeliness and uncertainty. Further, we develop a time-aware reweighting strategy with a teacher-student model to stabilize the adaptation process. Extensive experiments and ablation studies are conducted to verify the robustness and effectiveness of the proposed method. We believe this work will pave the way for thinking about adapting models into real-world applications by test-time adaptation algorithm.

\paragraph{Acknowledgements.} 
This paper was supported by National Key R\&D Program of China (No. 2021YFB3301503), and also supported by the National Natural Science Foundation of China under Grant No. 61902028.

\clearpage

{\small
\bibliographystyle{ieee_fullname}
\bibliography{reference}

\begin{thebibliography}{10}\itemsep=-1pt

\bibitem{AljundiLGB19}
Rahaf Aljundi, Min Lin, Baptiste Goujaud, and Yoshua Bengio.
\newblock Gradient based sample selection for online continual learning.
\newblock In {\em NeurIPS}, pages 11816--11825, 2019.

\bibitem{AzimiPRHB022}
Fatemeh Azimi, Sebastian Palacio, Federico Raue, J{\"{o}}rn Hees, Luca
  Bertinetto, and Andreas Dengel.
\newblock Self-supervised test-time adaptation on video data.
\newblock In {\em {WACV}}, pages 2603--2612, 2022.

\bibitem{BatesonLA22}
Mathilde Bateson, Herve Lombaert, and Ismail~Ben Ayed.
\newblock Test-time adaptation with shape moments for image segmentation.
\newblock In {\em MICCAI}, pages 736--745, 2022.

\bibitem{BlanchardLS11}
Gilles Blanchard, Gyemin Lee, and Clayton Scott.
\newblock Generalizing from several related classification tasks to a new
  unlabeled sample.
\newblock In {\em NeurIPS}, pages 2178--2186, 2011.

\bibitem{niid_boudiaf2022parameter}
Malik Boudiaf, Romain Mueller, Ismail Ben~Ayed, and Luca Bertinetto.
\newblock Parameter-free online test-time adaptation.
\newblock In {\em CVPR}, pages 8344--8353, 2022.

\bibitem{castro2018end}
Francisco~M Castro, Manuel~J Mar{\'\i}n-Jim{\'e}nez, Nicol{\'a}s Guil, Cordelia
  Schmid, and Karteek Alahari.
\newblock End-to-end incremental learning.
\newblock In {\em ECCV}, pages 233--248, 2018.

\bibitem{chen2022_contrastivetta}
Dian Chen, Dequan Wang, Trevor Darrell, and Sayna Ebrahimi.
\newblock Contrastive test-time adaptation.
\newblock In {\em CVPR}, pages 295--305, 2022.

\bibitem{chen2018domain}
Yuhua Chen, Wen Li, Christos Sakaridis, Dengxin Dai, and Luc Van~Gool.
\newblock Domain adaptive faster r-cnn for object detection in the wild.
\newblock In {\em CVPR}, pages 3339--3348, 2018.

\bibitem{ChiWYT21}
Zhixiang Chi, Yang Wang, Yuanhao Yu, and Jin Tang.
\newblock Test-time fast adaptation for dynamic scene deblurring via
  meta-auxiliary learning.
\newblock In {\em CVPR}, pages 9137--9146, 2021.

\bibitem{ChidlovskiiCC16}
Boris Chidlovskii, St{\'{e}}phane Clinchant, and Gabriela Csurka.
\newblock Domain adaptation in the absence of source domain data.
\newblock In {\em KDD}, pages 451--460, 2016.

\bibitem{ChoiYCY22}
Sungha Choi, Seunghan Yang, Seokeon Choi, and Sungrack Yun.
\newblock Improving test-time adaptation via shift-agnostic weight
  regularization and nearest source prototypes.
\newblock In {\em ECCV}, pages 440--458, 2022.

\bibitem{RobustBench}
Francesco Croce, Maksym Andriushchenko, Vikash Sehwag, Edoardo Debenedetti,
  Nicolas Flammarion, Mung Chiang, Prateek Mittal, and Matthias Hein.
\newblock Robustbench: a standardized adversarial robustness benchmark.
\newblock In {\em Neurips}, 2021.

\bibitem{DosovitskiyB0WZ21}
Alexey Dosovitskiy, Lucas Beyer, Alexander Kolesnikov, Dirk Weissenborn,
  Xiaohua Zhai, Thomas Unterthiner, Mostafa Dehghani, Matthias Minderer, Georg
  Heigold, Sylvain Gelly, Jakob Uszkoreit, and Neil Houlsby.
\newblock An image is worth 16x16 words: Transformers for image recognition at
  scale.
\newblock In {\em {ICLR}}, 2021.

\bibitem{DuXXQZS020}
Ying{-}Jun Du, Jun Xu, Huan Xiong, Qiang Qiu, Xiantong Zhen, Cees G.~M. Snoek,
  and Ling Shao.
\newblock Learning to learn with variational information bottleneck for domain
  generalization.
\newblock In {\em {ECCV}}, pages 200--216, 2020.

\bibitem{tta_document}
Sayna Ebrahimi, Sercan~{\"{O}}. Arik, and Tomas Pfister.
\newblock Test-time adaptation for visual document understanding.
\newblock {\em CoRR}, abs/2206.07240, 2022.

\bibitem{gandelsman2022testtime}
Yossi Gandelsman, Yu Sun, Xinlei Chen, and Alexei~A Efros.
\newblock Test-time training with masked autoencoders.
\newblock In {\em NeurIPS}, 2022.

\bibitem{GaninUAGLLML16}
Yaroslav Ganin, Evgeniya Ustinova, Hana Ajakan, Pascal Germain, Hugo
  Larochelle, Fran{\c{c}}ois Laviolette, Mario Marchand, and Victor~S.
  Lempitsky.
\newblock Domain-adversarial training of neural networks.
\newblock {\em J. Mach. Learn. Res.}, 17:59:1--59:35, 2016.

\bibitem{vpt_tta}
Yunhe Gao, Xingjian Shi, Yi Zhu, Hao Wang, Zhiqiang Tang, Xiong Zhou, Mu Li,
  and Dimitris~N. Metaxas.
\newblock Visual prompt tuning for test-time domain adaptation.
\newblock {\em CoRR}, abs/2210.04831, 2022.

\bibitem{note}
Taesik Gong, Jongheon Jeong, Taewon Kim, Yewon Kim, Jinwoo Shin, and Sung{-}Ju
  Lee.
\newblock Robust continual test-time adaptation: Instance-aware {BN} and
  prediction-balanced memory.
\newblock In {\em {NeurIPS}}, 2022.

\bibitem{goyal2022test}
Sachin Goyal, Mingjie Sun, Aditi Raghunathan, and J~Zico Kolter.
\newblock Test time adaptation via conjugate pseudo-labels.
\newblock In {\em NeurIPS}, 2022.

\bibitem{entropy_minimization}
Yves Grandvalet and Yoshua Bengio.
\newblock Semi-supervised learning by entropy minimization.
\newblock In {\em NeurIPS}, pages 529--536, 2004.

\bibitem{resnet}
Kaiming He, Xiangyu Zhang, Shaoqing Ren, and Jian Sun.
\newblock Deep residual learning for image recognition.
\newblock In {\em CVPR}, pages 770--778, 2016.

\bibitem{corruptions}
Dan Hendrycks and Thomas~G. Dietterich.
\newblock Benchmarking neural network robustness to common corruptions and
  perturbations.
\newblock In {\em ICLR}, 2019.

\bibitem{huang2022extrapolative}
Hengguan Huang, Xiangming Gu, Hao Wang, Chang Xiao, Hongfu Liu, and Ye Wang.
\newblock Extrapolative continuous-time bayesian neural network for fast
  training-free test-time adaptation.
\newblock In {\em NeurIPS}, 2022.

\bibitem{BN}
Sergey Ioffe and Christian Szegedy.
\newblock Batch normalization: Accelerating deep network training by reducing
  internal covariate shift.
\newblock In {\em {ICML}}, pages 448--456, 2015.

\bibitem{IwasawaM21}
Yusuke Iwasawa and Yutaka Matsuo.
\newblock Test-time classifier adjustment module for model-agnostic domain
  generalization.
\newblock In {\em NeurIPS}, pages 2427--2440, 2021.

\bibitem{jain2011online}
Vidit Jain and Erik Learned-Miller.
\newblock Online domain adaptation of a pre-trained cascade of classifiers.
\newblock In {\em CVPR}, pages 577--584, 2011.

\bibitem{tta_iclr1}
Minguk Jang and Sae{-}Young Chung.
\newblock Test-time adaptation via self-training with nearest neighbor
  information.
\newblock {\em CoRR}, abs/2207.10792, 2022.

\bibitem{KimH022}
Junho Kim, Inwoo Hwang, and Young~Min Kim.
\newblock Ev-tta: Test-time adaptation for event-based object recognition.
\newblock In {\em CVPR}, pages 17724--17733, 2022.

\bibitem{Adam}
Diederik~P. Kingma and Jimmy Ba.
\newblock Adam: {A} method for stochastic optimization.
\newblock In {\em ICLR}, 2015.

\bibitem{KirkpatrickPRVD16}
James Kirkpatrick, Razvan Pascanu, Neil~C. Rabinowitz, Joel Veness, Guillaume
  Desjardins, Andrei~A. Rusu, Kieran Milan, John Quan, Tiago Ramalho, Agnieszka
  Grabska{-}Barwinska, Demis Hassabis, Claudia Clopath, Dharshan Kumaran, and
  Raia Hadsell.
\newblock Overcoming catastrophic forgetting in neural networks.
\newblock {\em CoRR}, abs/1612.00796, 2016.

\bibitem{cifar}
Alex Krizhevsky, Geoffrey Hinton, et~al.
\newblock Learning multiple layers of features from tiny images.
\newblock 2009.

\bibitem{alexnet}
Alex Krizhevsky, Ilya Sutskever, and Geoffrey~E. Hinton.
\newblock Imagenet classification with deep convolutional neural networks.
\newblock In {\em NeurIPS}, pages 1097--1105, 2012.

\bibitem{kumar2020understanding}
Ananya Kumar, Tengyu Ma, and Percy Liang.
\newblock Understanding self-training for gradual domain adaptation.
\newblock In {\em ICML}, pages 5468--5479, 2020.

\bibitem{KunduVVB20}
Jogendra~Nath Kundu, Naveen Venkat, Rahul~M. V., and R.~Venkatesh Babu.
\newblock Universal source-free domain adaptation.
\newblock In {\em CVPR}, pages 4543--4552, 2020.

\bibitem{kurmi2021domain}
Vinod~K Kurmi, Venkatesh~K Subramanian, and Vinay~P Namboodiri.
\newblock Domain impression: A source data free domain adaptation method.
\newblock In {\em WACV}, pages 615--625, 2021.

\bibitem{LangeAMPJLST22}
Matthias~De Lange, Rahaf Aljundi, Marc Masana, Sarah Parisot, Xu Jia, Ales
  Leonardis, Gregory~G. Slabaugh, and Tinne Tuytelaars.
\newblock A continual learning survey: Defying forgetting in classification
  tasks.
\newblock {\em {IEEE} Trans. Pattern Anal. Mach. Intell.}, 44(7):3366--3385,
  2022.

\bibitem{DLeCunBH15}
Yann LeCun, Yoshua Bengio, and Geoffrey~E. Hinton.
\newblock Deep learning.
\newblock {\em Nat.}, 521(7553):436--444, 2015.

\bibitem{PL}
Dong-Hyun Lee et~al.
\newblock Pseudo-label: The simple and efficient semi-supervised learning
  method for deep neural networks.
\newblock In {\em Workshop on challenges in representation learning, ICML},
  volume~3, page 896, 2013.

\bibitem{LiYSH18}
Da Li, Yongxin Yang, Yi{-}Zhe Song, and Timothy~M. Hospedales.
\newblock Learning to generalize: Meta-learning for domain generalization.
\newblock In {\em AAAI}, pages 3490--3497, 2018.

\bibitem{LiPWK18}
Haoliang Li, Sinno~Jialin Pan, Shiqi Wang, and Alex~C. Kot.
\newblock Domain generalization with adversarial feature learning.
\newblock In {\em CVPR}, pages 5400--5409, 2018.

\bibitem{GDCAN}
Shuang Li, Binhui Xie, Qiuxia Lin, Chi~Harold Liu, Gao Huang, and Guoren Wang.
\newblock Generalized domain conditioned adaptation network.
\newblock {\em IEEE Trans. Pattern Anal. Mach. Intell.}, 44(8):4093--4109,
  2022.

\bibitem{TSA}
Shuang Li, Mixue Xie, Kaixiong Gong, Chi~Harold Liu, Yulin Wang, and Wei Li.
\newblock Transferable semantic augmentation for domain adaptation.
\newblock In {\em CVPR}, pages 11516--11525, 2021.

\bibitem{LiH18a}
Zhizhong Li and Derek Hoiem.
\newblock Learning without forgetting.
\newblock {\em {IEEE} Trans. Pattern Anal. Mach. Intell.}, 40(12):2935--2947,
  2018.

\bibitem{LiangHF20}
Jian Liang, Dapeng Hu, and Jiashi Feng.
\newblock Do we really need to access the source data? source hypothesis
  transfer for unsupervised domain adaptation.
\newblock In {\em {ICML}}, pages 6028--6039, 2020.

\bibitem{LiuKDBMA21}
Yuejiang Liu, Parth Kothari, Bastien van Delft, Baptiste Bellot{-}Gurlet,
  Taylor Mordan, and Alexandre Alahi.
\newblock {TTT++:} when does self-supervised test-time training fail or thrive?
\newblock In {\em NeurIPS}, pages 21808--21820, 2021.

\bibitem{liu2021source}
Yuang Liu, Wei Zhang, and Jun Wang.
\newblock Source-free domain adaptation for semantic segmentation.
\newblock In {\em CVPR}, pages 1215--1224, 2021.

\bibitem{DAN-PAMI}
Mingsheng Long, Yue Cao, Zhangjie Cao, Jianmin Wang, and Michael~I. Jordan.
\newblock Transferable representation learning with deep adaptation networks.
\newblock {\em {IEEE} Trans. Pattern Anal. Mach. Intell.}, 41(12):3071--3085,
  2019.

\bibitem{MaCZQZD22}
Wenao Ma, Cheng Chen, Shuang Zheng, Jing Qin, Huimao Zhang, and Qi Dou.
\newblock Test-time adaptation with calibration of medical image classification
  nets for label distribution shift.
\newblock In {\em {MICCAI}}, pages 313--323, 2022.

\bibitem{MahajanTS21}
Divyat Mahajan, Shruti Tople, and Amit Sharma.
\newblock Domain generalization using causal matching.
\newblock In {\em {ICML}}, pages 7313--7324, 2021.

\bibitem{MansourMR09}
Yishay Mansour, Mehryar Mohri, and Afshin Rostamizadeh.
\newblock Domain adaptation: Learning bounds and algorithms.
\newblock In {\em {COLT}}, 2009.

\bibitem{MuandetBS13}
Krikamol Muandet, David Balduzzi, and Bernhard Sch{\"{o}}lkopf.
\newblock Domain generalization via invariant feature representation.
\newblock In {\em ICML}, pages 10--18, 2013.

\bibitem{BN_Stat}
Zachary Nado, Shreyas Padhy, D. Sculley, Alexander D'Amour, Balaji
  Lakshminarayanan, and Jasper Snoek.
\newblock Evaluating prediction-time batch normalization for robustness under
  covariate shift.
\newblock {\em CoRR}, abs/2006.10963, 2020.

\bibitem{NiuW0CZZT22}
Shuaicheng Niu, Jiaxiang Wu, Yifan Zhang, Yaofo Chen, Shijian Zheng, Peilin
  Zhao, and Mingkui Tan.
\newblock Efficient test-time model adaptation without forgetting.
\newblock In {\em ICML}, pages 16888--16905, 2022.

\bibitem{EATA}
Shuaicheng Niu, Jiaxiang Wu, Yifan Zhang, Yaofo Chen, Shijian Zheng, Peilin
  Zhao, and Mingkui Tan.
\newblock Efficient test-time model adaptation without forgetting.
\newblock In {\em ICML}, volume 162, pages 16888--16905, 2022.

\bibitem{survey}
Sinno~Jialin Pan and Qiang Yang.
\newblock A survey on transfer learning.
\newblock {\em IEEE Trans. Knowl. Data Eng.}, 22(10):1345--1359, 2010.

\bibitem{paszke2019pytorch}
Adam Paszke, Sam Gross, Francisco Massa, Adam Lerer, James Bradbury, Gregory
  Chanan, Trevor Killeen, Zeming Lin, Natalia Gimelshein, Luca Antiga, et~al.
\newblock Pytorch: An imperative style, high-performance deep learning library.
\newblock In {\em NeurIPS}, pages 8024--8035, 2019.

\bibitem{DomainNet}
Xingchao Peng, Qinxun Bai, Xide Xia, Zijun Huang, Kate Saenko, and Bo Wang.
\newblock Moment matching for multi-source domain adaptation.
\newblock In {\em ICCV}, pages 1406--1415, 2019.

\bibitem{quinonero2008dataset}
Joaquin Quinonero-Candela, Masashi Sugiyama, Anton Schwaighofer, and Neil~D
  Lawrence.
\newblock {\em Dataset shift in machine learning}.
\newblock 2008.

\bibitem{RebuffiKSL17}
Sylvestre{-}Alvise Rebuffi, Alexander Kolesnikov, Georg Sperl, and Christoph~H.
  Lampert.
\newblock icarl: Incremental classifier and representation learning.
\newblock In {\em {CVPR}}, pages 5533--5542, 2017.

\bibitem{royer2015classifier}
Amelie Royer and Christoph~H Lampert.
\newblock Classifier adaptation at prediction time.
\newblock In {\em CVPR}, pages 1401--1409, 2015.

\bibitem{MCD}
Kuniaki Saito, Kohei Watanabe, Yoshitaka Ushiku, and Tatsuya Harada.
\newblock Maximum classifier discrepancy for unsupervised domain adaptation.
\newblock In {\em CVPR}, pages 3723--3732, 2018.

\bibitem{ShinTZSLGKY22}
Inkyu Shin, Yi{-}Hsuan Tsai, Bingbing Zhuang, Samuel Schulter, Buyu Liu, Sparsh
  Garg, In~So Kweon, and Kuk{-}Jin Yoon.
\newblock {MM-TTA:} multi-modal test-time adaptation for 3d semantic
  segmentation.
\newblock In {\em CVPR}, pages 16907--16916, 2022.

\bibitem{shu2022testtime}
Manli Shu, Weili Nie, De-An Huang, Zhiding Yu, Tom Goldstein, Anima Anandkumar,
  and Chaowei Xiao.
\newblock Test-time prompt tuning for zero-shot generalization in
  vision-language models.
\newblock In {\em NeurIPS}, 2022.

\bibitem{ttt_sun2020}
Yu Sun, Xiaolong Wang, Zhuang Liu, John Miller, Alexei Efros, and Moritz Hardt.
\newblock Test-time training with self-supervision for generalization under
  distribution shifts.
\newblock In {\em ICML}, pages 9229--9248, 2020.

\bibitem{TiwariKIS22}
Rishabh Tiwari, KrishnaTeja Killamsetty, Rishabh~K. Iyer, and Pradeep Shenoy.
\newblock {GCR:} gradient coreset based replay buffer selection for continual
  learning.
\newblock In {\em {CVPR}}, pages 99--108, 2022.

\bibitem{tsai2018learning}
Yi-Hsuan Tsai, Wei-Chih Hung, Samuel Schulter, Kihyuk Sohn, Ming-Hsuan Yang,
  and Manmohan Chandraker.
\newblock Learning to adapt structured output space for semantic segmentation.
\newblock In {\em CVPR}, pages 7472--7481, 2018.

\bibitem{Tzeng_2015_Simultaneous}
Eric Tzeng, Judy Hoffman, Trevor Darrell, and Kate Saenko.
\newblock Simultaneous deep transfer across domains and tasks.
\newblock In {\em {ICCV}}, pages 4068--4076, 2015.

\bibitem{Tzeng_ADDA}
Eric Tzeng, Judy Hoffman, Kate Saenko, and Trevor Darrell.
\newblock Adversarial discriminative domain adaptation.
\newblock In {\em {CVPR}}, pages 2962--2971, 2017.

\bibitem{tent_wang2020}
Dequan Wang, Evan Shelhamer, Shaoteng Liu, Bruno~A. Olshausen, and Trevor
  Darrell.
\newblock Tent: Fully test-time adaptation by entropy minimization.
\newblock In {\em {ICLR}}, 2021.

\bibitem{wang2022generalizing}
Jindong Wang, Cuiling Lan, Chang Liu, Yidong Ouyang, Tao Qin, Wang Lu, Yiqiang
  Chen, Wenjun Zeng, and Philip Yu.
\newblock Generalizing to unseen domains: A survey on domain generalization.
\newblock {\em IEEE Trans. Knowl. Data Eng.}, 2022.

\bibitem{wang2018deep}
Mei Wang and Weihong Deng.
\newblock Deep visual domain adaptation: A survey.
\newblock {\em Neurocomputing}, 312:135--153, 2018.

\bibitem{cotta}
Qin Wang, Olga Fink, Luc~Van Gool, and Dengxin Dai.
\newblock Continual test-time domain adaptation.
\newblock In {\em {CVPR}}, pages 7191--7201, 2022.

\bibitem{wulfmeier2018incremental}
Markus Wulfmeier, Alex Bewley, and Ingmar Posner.
\newblock Incremental adversarial domain adaptation for continually changing
  environments.
\newblock In {\em ICRA}, pages 4489--4495, 2018.

\bibitem{sepico}
Binhui Xie, Shuang Li, Mingjia Li, Chi~Harold Liu, Gao Huang, and Guoren Wang.
\newblock Sepico: Semantic-guided pixel contrast for domain adaptive semantic
  segmentation.
\newblock {\em IEEE Trans. Pattern Anal. Mach. Intell.}, pages 1--17, 2023.

\bibitem{ResNext}
Saining Xie, Ross Girshick, Piotr Doll{\'a}r, Zhuowen Tu, and Kaiming He.
\newblock Aggregated residual transformations for deep neural networks.
\newblock In {\em CVPR}, pages 5987--5995, 2017.

\bibitem{AFN2019Xu}
Ruijia Xu, Guanbin Li, Jihan Yang, and Liang Lin.
\newblock Larger norm more transferable: An adaptive feature norm approach for
  unsupervised domain adaptation.
\newblock In {\em {ICCV}}, pages 1426--1435, 2019.

\bibitem{XuLYRN21}
Zhenlin Xu, Deyi Liu, Junlin Yang, Colin Raffel, and Marc Niethammer.
\newblock Robust and generalizable visual representation learning via random
  convolutions.
\newblock In {\em {ICLR}}, 2021.

\bibitem{yang2021generalized}
Shiqi Yang, Yaxing Wang, Joost van~de Weijer, Luis Herranz, and Shangling Jui.
\newblock Generalized source-free domain adaptation.
\newblock In {\em ICCV}, pages 8978--8987, 2021.

\bibitem{wildresnet}
Sergey Zagoruyko and Nikos Komodakis.
\newblock Wide residual networks.
\newblock In {\em BMVC}, 2016.

\bibitem{zhang2022memo}
Marvin~Mengxin Zhang, Sergey Levine, and Chelsea Finn.
\newblock {MEMO}: Test time robustness via adaptation and augmentation.
\newblock In {\em NeurIPS}, 2022.

\bibitem{ZhangBCP22}
Yizhe Zhang, Shubhankar Borse, Hong Cai, and Fatih Porikli.
\newblock Auxadapt: Stable and efficient test-time adaptation for temporally
  consistent video semantic segmentation.
\newblock In {\em {WACV}}, pages 2633--2642, 2022.

\bibitem{zhou2022domain}
Kaiyang Zhou, Ziwei Liu, Yu Qiao, Tao Xiang, and Chen~Change Loy.
\newblock Domain generalization: A survey.
\newblock {\em {IEEE} Trans. Pattern Anal. Mach. Intell.}, 2022.

\bibitem{ZhouY0X21}
Kaiyang Zhou, Yongxin Yang, Yu Qiao, and Tao Xiang.
\newblock Domain generalization with mixstyle.
\newblock In {\em ICLR}, 2021.

\bibitem{zou2018unsupervised}
Yang Zou, Zhiding Yu, BVK~Vijaya Kumar, and Jinsong Wang.
\newblock Unsupervised domain adaptation for semantic segmentation via
  class-balanced self-training.
\newblock In {\em ECCV}, pages 289--305, 2018.

\end{thebibliography}
}

\clearpage

\section{Appendix}
\subsection{Discussion}
\paragraph{Societal impact.} \method enables adapting pre-trained models on continually changing distributions with correlatively sampled test streams without any more raw data or label requirements. Thus, our work may have a positive impact on communities to effectively deploy and adapt models in various real-world scenarios, which is economically and environmentally friendly. And since no training data is required, this protects data privacy and has potential commercial value. We carry out experiments on benchmark datasets and do not notice any societal issues. It does not involve sensitive attributes.

\paragraph{Future work.} Our work suggests a few promising directions for future
work. Firstly, the proposed \method is a preliminary attempt to perform test-time adaptation for the more realistic test stream under the setup \setting. One could experiment to improve the algorithm by replacing some parts of \method. More importantly, we hope that with this work, we can open a path to the original goal of test-time adaptation, which is performing test-time adaptation in real-world scenarios. Thus, one could improve \setting to make it more realistic.

\paragraph{Limitations.} \method achieves excellent performance on various tasks under the setup \setting as demonstrated in Section~{\color{red} 4} in the main paper, but we still find some limitations of it. Firstly, the adopted robust batch normalization (RBN) is a naive solution to the normalization of the correlatively sampled batch of data. This requires careful design of the value of $\alpha$ in RBN. 
Secondly, we observe that during the adaptation procedure of some methods like PL~\cite{PL} and TENT~\cite{tent_wang2020}, the model collapse finally. Although we design many strategies to stabilize the adaptation and model collapse never occurs in the experiments of \method, we are still missing a way to recover the model from the collapse state as a remedy.
Thirdly, category similarity is only one kind of correlation. Although we conduct experiments on different datasets with Dirichlet distribution to simulate correlatively sampled test streams, we still need to validate our approach in some real-world scenarios.

\subsection{Sensitivity to different hyper-parameters}
In this section, we conduct a detailed sensitivity analysis of the hyperparameters involved in \method. All experiments are conducted on CIFAR100$\to$CIFAR100-C, and the corruptions changes as {\it motion, snow, fog, shot, defocus, contrast, zoom, brightness, frost, elastic, glass, gaussian, pixelate, jpeg,} and {\it impulse}, and test streams are sampled correlatively with the Dirichlet parameter $\delta = 0.1$. When we investigate the sensitivity to a specific hyperparameter, other hyperparameters are fixed to the default values, i.e.,  $\lambda_t=1.0$, $\lambda_u=1.0$, $\alpha=0.05$, and $\nu=0.001$, for all experiments.
\begin{table}[h]
    \centering
    \caption{Classification error with different value of $\lambda_t / \lambda_u$.}
    \label{table:trade_off}
    \vspace{-2mm}
    \resizebox{\linewidth}{!}{
    \begin{tabular}{l|ccccccccc}
        \toprule
        $\lambda_t/\lambda_u$ & 0.0/2.0 & 0.5/1.5 & 1.0/1.0 & 1.5/ 0.5 & 2.0/ 0.0 \\
        \hline
        CIFAR100-C & 57.5 & 36.9 & \bf 35.0 & 35.9 & 38.9 \\
    \bottomrule
    \end{tabular}
    }
\end{table}

\paragraph{Trade-off between timeliness and uncertainty.}
When updating the memory bank, we take the timeliness and uncertainty of samples into account simultaneously, and $\lambda_t \text{ and } \lambda_u$ will make a trade-off between them. In Table~\ref{table:trade_off}, we show the results of \method with varying $\lambda_t / \lambda_u$, i.e., $\lambda_t / \lambda_u\in \{0.0/2.0, 0.5/1.5, 1.0/1.0, 1.5/0.5, 2.0/0.0 \}$. When we consider both of them, the results are relatively stable (35.0-36.9\%). When we only think about one side, the performance drops significantly. For example, when we set $\lambda_t / \lambda_u = 0.0/2.0$ which means only considering uncertainty, the performance drops 22.5\%. That's because some confident samples get stuck in the memory bank, making it not work the way we design it.

\begin{table}[h]
    \centering
    \caption{Classification error with varying $\alpha$}
    \label{table:alpha}
    \vspace{-2mm}
    \resizebox{\linewidth}{!}{
    \begin{tabular}{l|cccccc}
        \toprule
        $\alpha$ & 0.5 & 0.1 & 0.05& 0.01& 0.005 & 0.001 \\
        \hline
        CIFAR100-C & 39.0 & 36.0 & \bf 35.0 & 36.0 & 38.1 & 41.5 \\
    \bottomrule
    \end{tabular}
    }
\end{table}

\paragraph{Sensitivity to $\alpha$.} We show the results of \method with varying $\alpha$, i.e., $\alpha \in \{0.5, 0.1, 0.05, 0.01, 0.005, 0.001\}$ in Table~\ref{table:alpha}. A larger value of $\alpha$ means updating the global statistics faster and vice versa. We can see that \method achieves competitive results ($35.0-36.0\%$) at appropriate values of $\alpha$, i.e., $\alpha \in \{0.1, 0.05, 0.01\}$. Updating too aggressively or too gently can lead to unreliable estimates of statistics.

\begin{table}[h]
    \centering
    \caption{Classification error with varying $\nu$}
    \label{table:nu}
    \vspace{-2mm}
    \resizebox{\linewidth}{!}{
    \begin{tabular}{l|cccccc}
        \toprule
        $\nu$ & 0.05 & 0.01 & 0.005& 0.001& 0.0005 & 0.0001 \\
        \hline
        CIFAR100-C & 44.8 & 39.1 & 37.1 & \bf 35.0 & 37.6 & 43.6 \\
    \bottomrule
    \end{tabular}
    }
\end{table}

\paragraph{Sensitivity to $\nu$.} We show the results of \method with varying $\nu$, i.e., $\nu \in \{0.05, 0.01, 0.005, 0.001, 0.0005, 0.0001\}$ in Table~\ref{table:nu}. As we can see, the best performance is achieved at $\nu=0.001$. Updating the teacher model too quickly or too slowly can cause performance degradation.

\subsection{Additional experiment details and results}
\subsubsection{Compared methods}
\paragraph{BN}~\cite{BN_Stat} utilizes statistics of the current batch of data to normalize their feature maps without tuning any parameters.

\paragraph{PL}~\cite{PL} is based on BN~\cite{BN_Stat}, and adopts pseudo labels to train the affine parameters in BN layers.

\paragraph{TENT}~\cite{tent_wang2020} is the first to propose fully test-time adaptation. It adopts test-time batch normalization and utilizes entropy minimization to train the affine parameters of BN layers. We reimplement it following the released code \url{https://github.com/DequanWang/tent}.

\paragraph{LAME}~\cite{niid_boudiaf2022parameter} adapts the output of the pre-trained model by optimizing a group of latent variables without tuning any inner parts of the model. We reimplement it following the released code \url{https://github.com/fiveai/LAME}.

\paragraph{CoTTA}~\cite{cotta} considers performing test-time adaptation on continually changing distributions and propose augmentation-averaged pseudo-labels and stochastic restoration to address error accumulation and catastrophic forgetting. We reimplement it following the released code \url{https://github.com/qinenergy/cotta}.

\paragraph{NOTE}~\cite{note} proposes instance-aware normalization and prediction-balanced reservoir sampling to stable the adaptation on temporally correlated test streams. We reimplement it following the released code \url{https://github.com/TaesikGong/NOTE}.

\subsubsection{Simulate correlatively sampling} As we described in the scenarios of autonomous driving that the car will follow more vehicles on the highway or will encounter more pedestrians on the sidewalk, so we use the same category to simulate correlation. From a macro point of view, the test distribution $\distribution_{test}$ changes continually as $\distribution_0, \distribution_1, ...,\distribution_\infty$. During the period when $\distribution_{test} = \distribution_t$, we adopt Dirichlet distribution to simulate correlatively sampled test stream. More specifically, we consider dividing samples of $\mathcal{C}$ classes into $T$ slots. Firstly, we utilize Dirichlet distribution with parameter $\gamma$ to generate the partition criterion $q\in\mathbb{R}^{\mathcal{C}\times T}$. Then for each class $c$, we split samples into $T$ parts according to $q_c$ and assign each part to each slot respectively. Finally, we concatenate all slots to simulate the correlatively sampled test stream for $\distribution_{test} = \distribution_t$. And as $\distribution_{test}$ changes, we use the above method again to generate the test stream.

\subsubsection{Detailed results of different orders}
We report the average classification error of ten different distribution changing orders in Table~{\color{red} 6} of the main paper. And then we present the specific results here, including Table~\ref{table:cifar10_order0}, \ref{table:cifar10_order1}, \ref{table:cifar10_order2}, \ref{table:cifar10_order3}, \ref{table:cifar10_order4}, \ref{table:cifar10_order5}, \ref{table:cifar10_order6}, \ref{table:cifar10_order7}, \ref{table:cifar10_order8}, and \ref{table:cifar10_order9} for CIFAR10$\to$CIFAR10-C and Table~\ref{table:cifar100_order0}, \ref{table:cifar100_order1}, \ref{table:cifar100_order2}, \ref{table:cifar100_order3}, \ref{table:cifar100_order4}, \ref{table:cifar100_order5}, \ref{table:cifar100_order6}, \ref{table:cifar100_order7}, \ref{table:cifar100_order8}, and \ref{table:cifar100_order9} for CIFAR100$\to$CIFAR100-C. We can see consistently superior performance of \method.
One thing to mention is that on DomainNet we use alphabetical order to determine the order of domain changes.

\begin{table*}[htb]
	\centering
	\caption{Average classification error of the task CIFAR10 $\to$ CIFAR10-C while continually adapting to different corruptions at the highest severity 5 with correlatively sampled test stream under the proposed setup \setting.}
	\label{table:cifar10_order0}
	\VspaceBefore
	\resizebox{\textwidth}{!}
	{
		\renewcommand{\arraystretch}{0.8}
		{
\begin{tabular}{l|ccccccccccccccc|c}

				\toprule[1.2pt]

				Time & \multicolumn{15}{l|}{$t\xrightarrow{\hspace*{18.5cm}}$}& \\ \hline

				Method & \rotatebox[origin=c]{45}{brightness} & \rotatebox[origin=c]{45}{pixelate} & \rotatebox[origin=c]{45}{gaussian} & \rotatebox[origin=c]{45}{motion} & \rotatebox[origin=c]{45}{zoom} & \rotatebox[origin=c]{45}{glass} & \rotatebox[origin=c]{45}{impulse} & \rotatebox[origin=c]{45}{jpeg} & \rotatebox[origin=c]{45}{defocus} & \rotatebox[origin=c]{45}{elastic} & \rotatebox[origin=c]{45}{shot} & \rotatebox[origin=c]{45}{frost} & \rotatebox[origin=c]{45}{snow} & \rotatebox[origin=c]{45}{fog} & \rotatebox[origin=c]{45}{contrast}  & Avg. \\

				\midrule

				Source & \underline{9.3} & 58.5 & 72.3 & 34.8 & 42.0 & 54.3 & 72.9 & 30.3 & 46.9 & \underline{26.6} & 65.7 & 41.3 & 25.1 & 26.0 & 46.7 & 43.5 \\

				BN~\cite{BN_Stat} & 71.1 & 75.2 & 76.8 & 74.2 & 73.7 & 80.1 & 79.3 & 77.5 & 73.8 & 77.7 & 77.2 & 73.3 & 73.8 & 72.7 & 71.7 & 75.2 \\

				PL~\cite{PL} & 71.7 & 75.9 & 80.2 & 78.4 & 80.2 & 85.2 & 85.3 & 85.4 & 85.1 & 86.7 & 87.9 & 87.9 & 88.1 & 88.3 & 87.9 & 83.6 \\

				TENT~\cite{tent_wang2020} & 71.6 & 75.9 & 81.3 & 80.5 & 82.3 & 85.6 & 87.1 & 87.0 & 87.1 & 88.1 & 88.2 & 87.8 & 87.9 & 88.3 & 88.2 & 84.4 \\

				LAME~\cite{niid_boudiaf2022parameter} & \bf 5.4 & 56.8 & 73.1 & 29.1 & 37.0 & 50.5 & 71.4 & \bf 22.3 & 42.8 & \bf 18.6 & 65.5 & 37.3 & \bf 18.8 & \underline{20.4} & 43.6 & 39.5 \\

				CoTTA~\cite{cotta} & 75.0 & 79.8 & 83.1 & 83.4 & 83.2 & 84.0 & 84.5 & 83.2 & 83.5 & 83.3 & 83.6 & 83.0 & 83.0 & 83.4 & 83.7 & 82.6 \\
\end{tabular}
		}
	}
	\VspaceAfter
\end{table*}

\begin{table*}[htb]
	\centering
	\caption{Average classification error of the task CIFAR10 $\to$ CIFAR10-C while continually adapting to different corruptions at the highest severity 5 with correlatively sampled test stream under the proposed setup \setting.}
	\label{table:cifar10_order1}
	\VspaceBefore
	\resizebox{\textwidth}{!}
	{
		\renewcommand{\arraystretch}{0.8}
		{
			\begin{tabular}{l|ccccccccccccccc|c}

				\toprule[1.2pt]

				Time & \multicolumn{15}{l|}{$t\xrightarrow{\hspace*{18.5cm}}$}& \\ \hline

				Method & \rotatebox[origin=c]{45}{jpeg} & \rotatebox[origin=c]{45}{shot} & \rotatebox[origin=c]{45}{zoom} & \rotatebox[origin=c]{45}{frost} & \rotatebox[origin=c]{45}{contrast} & \rotatebox[origin=c]{45}{fog} & \rotatebox[origin=c]{45}{defocus} & \rotatebox[origin=c]{45}{elastic} & \rotatebox[origin=c]{45}{gaussian} & \rotatebox[origin=c]{45}{brightness} & \rotatebox[origin=c]{45}{glass} & \rotatebox[origin=c]{45}{impulse} & \rotatebox[origin=c]{45}{pixelate} & \rotatebox[origin=c]{45}{snow} & \rotatebox[origin=c]{45}{motion}  & Avg. \\

				\midrule

				Source & \underline{30.3} & 65.7 & 42.0 & 41.3 & 46.7 & 26.0 & 46.9 & \underline{26.6} & 72.3 & \underline{9.3} & 54.3 & 72.9 & 58.5 & 25.1 & 34.8 & 43.5 \\

				BN~\cite{BN_Stat} & 77.6 & 75.8 & 73.4 & 74.1 & 73.1 & 72.5 & 72.9 & 77.1 & 77.2 & 72.2 & 79.9 & 79.9 & 75.5 & 74.6 & 72.9 & 75.2 \\

				PL~\cite{PL} & 77.6 & 77.1 & 76.6 & 78.3 & 77.5 & 79.8 & 82.0 & 84.8 & 86.1 & 83.5 & 87.8 & 87.1 & 86.5 & 85.6 & 85.7 & 82.4 \\

				TENT~\cite{tent_wang2020} & 78.5 & 78.2 & 79.2 & 81.8 & 84.8 & 84.8 & 86.4 & 87.3 & 87.9 & 86.7 & 87.3 & 87.8 & 87.2 & 87.5 & 87.1 & 84.8 \\

				LAME~\cite{niid_boudiaf2022parameter} & \bf 22.5 & 65.2 & 37.0 & 37.1 & 44.0 & \bf 20.3 & 41.7 & \bf 18.7 & 72.8 & \bf 5.2 & 51.2 & 71.5 & 57.0 & \bf 19.0 & 29.4 & 39.5 \\

				CoTTA~\cite{cotta} & 78.5 & 81.0 & 82.8 & 84.1 & 84.9 & 83.4 & 83.5 & 83.5 & 84.5 & 83.3 & 84.7 & 84.6 & 83.0 & 84.4 & 83.4 & 83.3 \\
\end{tabular}
		}
	}
	\VspaceAfter
\end{table*}

\begin{table*}[htb]
	\centering
	\caption{Average classification error of the task CIFAR10 $\to$ CIFAR10-C while continually adapting to different corruptions at the highest severity 5 with correlatively sampled test stream under the proposed setup \setting.}
	\label{table:cifar10_order2}
	\VspaceBefore
	\resizebox{\textwidth}{!}
	{
		\renewcommand{\arraystretch}{0.8}
		{
			\begin{tabular}{l|ccccccccccccccc|c}

				\toprule[1.2pt]

				Time & \multicolumn{15}{l|}{$t\xrightarrow{\hspace*{18.5cm}}$}& \\ \hline

				Method & \rotatebox[origin=c]{45}{contrast} & \rotatebox[origin=c]{45}{defocus} & \rotatebox[origin=c]{45}{gaussian} & \rotatebox[origin=c]{45}{shot} & \rotatebox[origin=c]{45}{snow} & \rotatebox[origin=c]{45}{frost} & \rotatebox[origin=c]{45}{glass} & \rotatebox[origin=c]{45}{zoom} & \rotatebox[origin=c]{45}{elastic} & \rotatebox[origin=c]{45}{jpeg} & \rotatebox[origin=c]{45}{pixelate} & \rotatebox[origin=c]{45}{brightness} & \rotatebox[origin=c]{45}{impulse} & \rotatebox[origin=c]{45}{motion} & \rotatebox[origin=c]{45}{fog}  & Avg. \\

				\midrule

				Source & 46.7 & 46.9 & 72.3 & 65.7 & 25.1 & 41.3 & 54.3 & 42.0 & \underline{26.6} & 30.3 & 58.5 & \underline{9.3} & 72.9 & 34.8 & 26.0 & 43.5 \\

				BN~\cite{BN_Stat} & 72.3 & 72.6 & 76.9 & 77.1 & 74.8 & 73.5 & 80.0 & 73.2 & 77.4 & 78.6 & 76.4 & 71.0 & 79.1 & 73.9 & 71.5 & 75.2 \\

				PL~\cite{PL} & 72.4 & 75.3 & 80.7 & 82.6 & 83.3 & 83.5 & 86.6 & 85.7 & 86.6 & 88.4 & 87.5 & 86.6 & 88.3 & 88.2 & 86.8 & 84.1 \\

				TENT~\cite{tent_wang2020} & 73.5 & 77.9 & 85.5 & 86.9 & 87.6 & 87.8 & 88.3 & 87.7 & 88.6 & 89.2 & 88.5 & 88.5 & 89.3 & 88.6 & 88.6 & 86.4 \\

				LAME~\cite{niid_boudiaf2022parameter} & 43.5 & 42.3 & 73.1 & 65.3 & \bf 19.2 & 37.3 & 51.1 & 36.8 & \bf 18.5 & \bf 22.5 & 56.9 & \bf 5.5 & 71.1 & 29.1 & \underline{20.5} & 39.5 \\

				CoTTA~\cite{cotta} & 79.4 & 80.3 & 83.8 & 83.9 & 83.9 & 83.4 & 85.0 & 83.2 & 85.1 & 84.3 & 83.9 & 83.3 & 84.7 & 83.9 & 82.5 & 83.4 \\
\end{tabular}
		}
	}
	\VspaceAfter
\end{table*}

\begin{table*}[htb]
	\centering
	\caption{Average classification error of the task CIFAR10 $\to$ CIFAR10-C while continually adapting to different corruptions at the highest severity 5 with correlatively sampled test stream under the proposed setup \setting.}
	\label{table:cifar10_order3}
	\VspaceBefore
	\resizebox{\textwidth}{!}
	{
		\renewcommand{\arraystretch}{0.8}
		{
			\begin{tabular}{l|ccccccccccccccc|c}

				\toprule[1.2pt]

				Time & \multicolumn{15}{l|}{$t\xrightarrow{\hspace*{18.5cm}}$}& \\ \hline

				Method & \rotatebox[origin=c]{45}{shot} & \rotatebox[origin=c]{45}{fog} & \rotatebox[origin=c]{45}{glass} & \rotatebox[origin=c]{45}{pixelate} & \rotatebox[origin=c]{45}{snow} & \rotatebox[origin=c]{45}{elastic} & \rotatebox[origin=c]{45}{brightness} & \rotatebox[origin=c]{45}{impulse} & \rotatebox[origin=c]{45}{defocus} & \rotatebox[origin=c]{45}{frost} & \rotatebox[origin=c]{45}{contrast} & \rotatebox[origin=c]{45}{gaussian} & \rotatebox[origin=c]{45}{motion} & \rotatebox[origin=c]{45}{jpeg} & \rotatebox[origin=c]{45}{zoom}  & Avg. \\

				\midrule

				Source & 65.7 & 26.0 & 54.3 & 58.5 & 25.1 & \underline{26.6} & \underline{9.3} & 72.9 & 46.9 & 41.3 & 46.7 & 72.3 & 34.8 & \underline{30.3} & 42.0 & 43.5 \\

				BN~\cite{BN_Stat} & 76.4 & 72.0 & 80.4 & 76.2 & 74.8 & 77.0 & 71.1 & 79.6 & 73.8 & 74.4 & 73.0 & 77.0 & 72.5 & 78.3 & 72.5 & 75.3 \\

				PL~\cite{PL} & 77.0 & 73.3 & 82.4 & 79.8 & 81.0 & 82.3 & 79.5 & 84.4 & 82.7 & 83.5 & 83.5 & 85.5 & 84.8 & 87.0 & 84.5 & 82.1 \\

				TENT~\cite{tent_wang2020} & 76.9 & 74.6 & 82.3 & 81.7 & 82.0 & 84.9 & 84.8 & 87.3 & 86.6 & 87.3 & 87.6 & 89.2 & 88.3 & 88.9 & 87.3 & 84.6 \\

				LAME~\cite{niid_boudiaf2022parameter} & 65.3 & \bf 20.6 & 50.9 & 56.7 & \bf 19.2 & \bf 18.8 & \bf 5.4 & 71.8 & 42.8 & 37.2 & 43.3 & 73.2 & 29.4 & \bf 22.6 & 36.9 & 39.6 \\

				CoTTA~\cite{cotta} & 77.4 & 77.6 & 83.8 & 81.9 & 82.2 & 82.6 & 80.4 & 83.3 & 82.3 & 81.5 & 82.7 & 82.6 & 81.1 & 82.9 & 81.0 & 81.6 \\
\end{tabular}
		}
	}
	\VspaceAfter
\end{table*}

\begin{table*}[htb]
	\centering
	\caption{Average classification error of the task CIFAR10 $\to$ CIFAR10-C while continually adapting to different corruptions at the highest severity 5 with correlatively sampled test stream under the proposed setup \setting.}
	\label{table:cifar10_order4}
	\VspaceBefore
	\resizebox{\textwidth}{!}
	{
		\renewcommand{\arraystretch}{0.8}
		{
			\begin{tabular}{l|ccccccccccccccc|c}

				\toprule[1.2pt]

				Time & \multicolumn{15}{l|}{$t\xrightarrow{\hspace*{18.5cm}}$}& \\ \hline

				Method & \rotatebox[origin=c]{45}{pixelate} & \rotatebox[origin=c]{45}{glass} & \rotatebox[origin=c]{45}{zoom} & \rotatebox[origin=c]{45}{snow} & \rotatebox[origin=c]{45}{fog} & \rotatebox[origin=c]{45}{impulse} & \rotatebox[origin=c]{45}{brightness} & \rotatebox[origin=c]{45}{motion} & \rotatebox[origin=c]{45}{frost} & \rotatebox[origin=c]{45}{jpeg} & \rotatebox[origin=c]{45}{gaussian} & \rotatebox[origin=c]{45}{shot} & \rotatebox[origin=c]{45}{contrast} & \rotatebox[origin=c]{45}{defocus} & \rotatebox[origin=c]{45}{elastic}  & Avg. \\

				\midrule

				Source & 58.5 & 54.3 & 42.0 & 25.1 & 26.0 & 72.9 & \underline{9.3} & 34.8 & 41.3 & \underline{30.3} & 72.3 & 65.7 & 46.7 & 46.9 & \underline{26.6} & 43.5 \\

				BN~\cite{BN_Stat} & 76.0 & 79.6 & 73.3 & 75.2 & 72.9 & 79.8 & 71.1 & 73.5 & 74.1 & 78.6 & 77.4 & 76.1 & 72.0 & 73.8 & 76.4 & 75.3 \\

				PL~\cite{PL} & 76.7 & 81.3 & 77.4 & 80.3 & 81.2 & 86.3 & 83.3 & 85.9 & 86.2 & 87.7 & 88.1 & 88.4 & 87.4 & 87.6 & 87.7 & 84.4 \\

				TENT~\cite{tent_wang2020} & 76.4 & 80.2 & 77.8 & 81.2 & 83.0 & 87.1 & 85.6 & 87.2 & 87.6 & 88.7 & 88.6 & 88.9 & 88.5 & 88.6 & 88.2 & 85.2 \\

				LAME~\cite{niid_boudiaf2022parameter} & 56.9 & \underline{50.7} & 37.0 & \bf 19.0 & \underline{20.3} & 71.5 & \bf 5.4 & 29.2 & 37.2 & \bf 22.5 & 73.0 & 65.3 & 43.8 & 42.4 & \bf 18.7 & 39.5 \\

				CoTTA~\cite{cotta} & 77.1 & 83.6 & 84.1 & 84.8 & 84.4 & 85.2 & 84.0 & 84.3 & 84.9 & 84.9 & 85.0 & 84.7 & 85.3 & 84.4 & 84.3 & 84.1 \\
\end{tabular}
		}
	}
	\VspaceAfter
\end{table*}

\begin{table*}[htb]
	\centering
	\caption{Average classification error of the task CIFAR10 $\to$ CIFAR10-C while continually adapting to different corruptions at the highest severity 5 with correlatively sampled test stream under the proposed setup \setting.}
	\label{table:cifar10_order5}
	\VspaceBefore
	\resizebox{\textwidth}{!}
	{
		\renewcommand{\arraystretch}{0.8}
		{
			\begin{tabular}{l|ccccccccccccccc|c}

				\toprule[1.2pt]

				Time & \multicolumn{15}{l|}{$t\xrightarrow{\hspace*{18.5cm}}$}& \\ \hline

				Method & \rotatebox[origin=c]{45}{motion} & \rotatebox[origin=c]{45}{snow} & \rotatebox[origin=c]{45}{fog} & \rotatebox[origin=c]{45}{shot} & \rotatebox[origin=c]{45}{defocus} & \rotatebox[origin=c]{45}{contrast} & \rotatebox[origin=c]{45}{zoom} & \rotatebox[origin=c]{45}{brightness} & \rotatebox[origin=c]{45}{frost} & \rotatebox[origin=c]{45}{elastic} & \rotatebox[origin=c]{45}{glass} & \rotatebox[origin=c]{45}{gaussian} & \rotatebox[origin=c]{45}{pixelate} & \rotatebox[origin=c]{45}{jpeg} & \rotatebox[origin=c]{45}{impulse}  & Avg. \\

				\midrule

				Source & 34.8 & 25.1 & 26.0 & 65.7 & 46.9 & 46.7 & 42.0 & \underline{9.3} & 41.3 & \underline{26.6} & 54.3 & 72.3 & 58.5 & 30.3 & 72.9 & 43.5 \\

				BN~\cite{BN_Stat} & 73.2 & 73.4 & 72.7 & 77.2 & 73.7 & 72.5 & 72.9 & 71.0 & 74.1 & 77.7 & 80.0 & 76.9 & 75.5 & 78.3 & 79.0 & 75.2 \\

				PL~\cite{PL} & 73.9 & 75.0 & 75.6 & 81.0 & 79.9 & 80.6 & 82.0 & 83.2 & 85.3 & 87.3 & 88.3 & 87.5 & 87.5 & 87.5 & 88.2 & 82.9 \\

				TENT~\cite{tent_wang2020} & 74.3 & 77.4 & 80.1 & 86.2 & 86.7 & 87.3 & 87.9 & 87.4 & 88.2 & 89.0 & 89.2 & 89.0 & 88.3 & 89.7 & 89.2 & 86.0 \\

				LAME~\cite{niid_boudiaf2022parameter} & 29.5 & \bf 19.0 & \underline{20.3} & 65.3 & 42.4 & 43.4 & 36.8 & \bf 5.4 & 37.2 & \bf 18.6 & 51.2 & 73.2 & 57.0 & \bf 22.6 & 71.3 & 39.5 \\

				CoTTA~\cite{cotta} & 77.1 & 80.6 & 83.1 & 84.4 & 83.9 & 84.2 & 83.1 & 82.6 & 84.4 & 84.2 & 84.5 & 84.6 & 82.7 & 83.8 & 84.9 & 83.2 \\
\end{tabular}
		}
	}
	\VspaceAfter
\end{table*}

\begin{table*}[htb]
	\centering
	\caption{Average classification error of the task CIFAR10 $\to$ CIFAR10-C while continually adapting to different corruptions at the highest severity 5 with correlatively sampled test stream under the proposed setup \setting.}
	\label{table:cifar10_order6}
	\VspaceBefore
	\resizebox{\textwidth}{!}
	{
		\renewcommand{\arraystretch}{0.8}
		{
			\begin{tabular}{l|ccccccccccccccc|c}

				\toprule[1.2pt]

				Time & \multicolumn{15}{l|}{$t\xrightarrow{\hspace*{18.5cm}}$}& \\ \hline

				Method & \rotatebox[origin=c]{45}{frost} & \rotatebox[origin=c]{45}{impulse} & \rotatebox[origin=c]{45}{jpeg} & \rotatebox[origin=c]{45}{contrast} & \rotatebox[origin=c]{45}{zoom} & \rotatebox[origin=c]{45}{glass} & \rotatebox[origin=c]{45}{pixelate} & \rotatebox[origin=c]{45}{snow} & \rotatebox[origin=c]{45}{defocus} & \rotatebox[origin=c]{45}{motion} & \rotatebox[origin=c]{45}{brightness} & \rotatebox[origin=c]{45}{elastic} & \rotatebox[origin=c]{45}{shot} & \rotatebox[origin=c]{45}{fog} & \rotatebox[origin=c]{45}{gaussian}  & Avg. \\

				\midrule

				Source & 41.3 & 72.9 & \underline{30.3} & 46.7 & 42.0 & 54.3 & 58.5 & 25.1 & 46.9 & 34.8 & \underline{9.3} & \underline{26.6} & 65.7 & 26.0 & 72.3 & 43.5 \\

				BN~\cite{BN_Stat} & 73.8 & 79.1 & 77.9 & 73.0 & 73.7 & 80.1 & 75.7 & 74.4 & 73.7 & 74.0 & 71.7 & 77.0 & 75.9 & 72.8 & 76.2 & 75.3 \\

				PL~\cite{PL} & 74.2 & 80.9 & 80.4 & 79.5 & 81.8 & 85.9 & 83.9 & 85.1 & 84.7 & 85.9 & 85.9 & 86.7 & 87.2 & 87.0 & 87.8 & 83.8 \\

				TENT~\cite{tent_wang2020} & 73.9 & 80.3 & 81.8 & 81.6 & 83.6 & 86.3 & 85.6 & 85.7 & 86.4 & 87.7 & 87.4 & 88.8 & 88.8 & 88.5 & 88.4 & 85.0 \\

				LAME~\cite{niid_boudiaf2022parameter} & 37.4 & 71.8 & \bf 22.4 & 43.5 & 37.0 & 50.5 & 57.0 & \bf 19.0 & 42.8 & 29.1 & \bf 5.4 & \bf 18.7 & 65.2 & \underline{20.4} & 72.9 & 39.5 \\

				CoTTA~\cite{cotta} & 76.5 & 82.2 & 82.8 & 85.0 & 82.9 & 85.0 & 83.0 & 82.9 & 83.5 & 83.4 & 82.6 & 83.7 & 83.2 & 83.3 & 83.6 & 82.9 \\
\end{tabular}
		}
	}
	\VspaceAfter
\end{table*}

\begin{table*}[htb]
	\centering
	\caption{Average classification error of the task CIFAR10 $\to$ CIFAR10-C while continually adapting to different corruptions at the highest severity 5 with correlatively sampled test stream under the proposed setup \setting.}
	\label{table:cifar10_order7}
	\VspaceBefore
	\resizebox{\textwidth}{!}
	{
		\renewcommand{\arraystretch}{0.8}
		{
			\begin{tabular}{l|ccccccccccccccc|c}

				\toprule[1.2pt]

				Time & \multicolumn{15}{l|}{$t\xrightarrow{\hspace*{18.5cm}}$}& \\ \hline

				Method & \rotatebox[origin=c]{45}{defocus} & \rotatebox[origin=c]{45}{motion} & \rotatebox[origin=c]{45}{zoom} & \rotatebox[origin=c]{45}{shot} & \rotatebox[origin=c]{45}{gaussian} & \rotatebox[origin=c]{45}{glass} & \rotatebox[origin=c]{45}{jpeg} & \rotatebox[origin=c]{45}{fog} & \rotatebox[origin=c]{45}{contrast} & \rotatebox[origin=c]{45}{pixelate} & \rotatebox[origin=c]{45}{frost} & \rotatebox[origin=c]{45}{snow} & \rotatebox[origin=c]{45}{brightness} & \rotatebox[origin=c]{45}{elastic} & \rotatebox[origin=c]{45}{impulse}  & Avg. \\

				\midrule

				Source & 46.9 & 34.8 & 42.0 & 65.7 & 72.3 & 54.3 & 30.3 & 26.0 & 46.7 & 58.5 & 41.3 & 25.1 & \underline{9.3} & \underline{26.6} & 72.9 & 43.5 \\

				BN~\cite{BN_Stat} & 72.8 & 72.7 & 73.3 & 77.2 & 77.3 & 80.0 & 77.6 & 72.6 & 73.3 & 76.6 & 73.8 & 74.1 & 70.3 & 77.5 & 79.0 & 75.2 \\

				PL~\cite{PL} & 73.2 & 74.6 & 76.5 & 81.7 & 82.8 & 84.6 & 85.1 & 84.6 & 86.2 & 86.4 & 86.1 & 87.1 & 86.8 & 88.4 & 88.1 & 83.5 \\

				TENT~\cite{tent_wang2020} & 73.7 & 74.3 & 77.1 & 82.5 & 84.3 & 86.9 & 87.4 & 86.6 & 88.0 & 88.5 & 88.1 & 88.5 & 88.4 & 89.4 & 88.9 & 84.8 \\

				LAME~\cite{niid_boudiaf2022parameter} & 42.5 & 29.3 & 37.0 & 65.3 & 73.2 & 50.5 & \bf 22.5 & \underline{20.5} & 43.5 & 56.9 & 37.1 & \bf 18.9 & \bf 5.4 & \bf 18.5 & 71.3 & 39.5 \\

				CoTTA~\cite{cotta} & 76.3 & 79.8 & 82.4 & 83.3 & 83.8 & 84.5 & 83.1 & 82.7 & 84.7 & 82.9 & 83.0 & 83.3 & 81.4 & 83.8 & 83.8 & 82.6 \\
\end{tabular}
		}
	}
	\VspaceAfter
\end{table*}

\begin{table*}[htb]
	\centering
	\caption{Average classification error of the task CIFAR10 $\to$ CIFAR10-C while continually adapting to different corruptions at the highest severity 5 with correlatively sampled test stream under the proposed setup \setting.}
	\label{table:cifar10_order8}
	\VspaceBefore
	\resizebox{\textwidth}{!}
	{
		\renewcommand{\arraystretch}{0.8}
		{
			\begin{tabular}{l|ccccccccccccccc|c}

				\toprule[1.2pt]

				Time & \multicolumn{15}{l|}{$t\xrightarrow{\hspace*{18.5cm}}$}& \\ \hline

				Method & \rotatebox[origin=c]{45}{glass} & \rotatebox[origin=c]{45}{zoom} & \rotatebox[origin=c]{45}{impulse} & \rotatebox[origin=c]{45}{fog} & \rotatebox[origin=c]{45}{snow} & \rotatebox[origin=c]{45}{jpeg} & \rotatebox[origin=c]{45}{gaussian} & \rotatebox[origin=c]{45}{frost} & \rotatebox[origin=c]{45}{shot} & \rotatebox[origin=c]{45}{brightness} & \rotatebox[origin=c]{45}{contrast} & \rotatebox[origin=c]{45}{motion} & \rotatebox[origin=c]{45}{pixelate} & \rotatebox[origin=c]{45}{defocus} & \rotatebox[origin=c]{45}{elastic}  & Avg. \\

				\midrule

				Source & 54.3 & 42.0 & 72.9 & 26.0 & 25.1 & \underline{30.3} & 72.3 & 41.3 & 65.7 & \underline{9.3} & 46.7 & 34.8 & 58.5 & 46.9 & \underline{26.6} & 43.5 \\

				BN~\cite{BN_Stat} & 79.7 & 72.3 & 79.8 & 73.2 & 74.7 & 77.7 & 76.6 & 73.2 & 77.1 & 72.2 & 73.0 & 73.3 & 75.5 & 73.8 & 76.4 & 75.2 \\

				PL~\cite{PL} & 79.6 & 73.2 & 81.3 & 77.3 & 79.1 & 83.0 & 83.2 & 83.0 & 85.5 & 84.3 & 87.0 & 86.9 & 86.4 & 86.5 & 87.6 & 82.9 \\

				TENT~\cite{tent_wang2020} & 79.5 & 74.1 & 84.2 & 82.2 & 84.5 & 86.5 & 86.7 & 85.9 & 87.2 & 86.6 & 86.8 & 87.3 & 86.9 & 87.4 & 87.3 & 84.9 \\

				LAME~\cite{niid_boudiaf2022parameter} & 50.8 & 36.9 & 71.3 & \bf 20.6 & \bf 19.2 & \bf 22.4 & 72.5 & 37.2 & 65.4 & \bf 5.2 & 43.3 & 29.1 & 57.0 & 42.4 & \bf 18.7 & 39.5 \\

				CoTTA~\cite{cotta} & 81.5 & 79.4 & 85.2 & 84.1 & 84.5 & 84.2 & 84.8 & 84.0 & 84.8 & 83.2 & 85.2 & 83.8 & 83.2 & 84.6 & 83.6 & 83.7 \\
\end{tabular}
		}
	}
	\VspaceAfter
\end{table*}

\begin{table*}[htb]
	\centering
	\caption{Average classification error of the task CIFAR10 $\to$ CIFAR10-C while continually adapting to different corruptions at the highest severity 5 with correlatively sampled test stream under the proposed setup \setting.}
	\label{table:cifar10_order9}
	\VspaceBefore
	\resizebox{\textwidth}{!}
	{
		\renewcommand{\arraystretch}{0.8}
		{
			\begin{tabular}{l|ccccccccccccccc|c}

				\toprule[1.2pt]

				Time & \multicolumn{15}{l|}{$t\xrightarrow{\hspace*{18.5cm}}$}& \\ \hline

				Method & \rotatebox[origin=c]{45}{contrast} & \rotatebox[origin=c]{45}{gaussian} & \rotatebox[origin=c]{45}{defocus} & \rotatebox[origin=c]{45}{zoom} & \rotatebox[origin=c]{45}{frost} & \rotatebox[origin=c]{45}{glass} & \rotatebox[origin=c]{45}{jpeg} & \rotatebox[origin=c]{45}{fog} & \rotatebox[origin=c]{45}{pixelate} & \rotatebox[origin=c]{45}{elastic} & \rotatebox[origin=c]{45}{shot} & \rotatebox[origin=c]{45}{impulse} & \rotatebox[origin=c]{45}{snow} & \rotatebox[origin=c]{45}{motion} & \rotatebox[origin=c]{45}{brightness}  & Avg. \\

				\midrule

				Source & 46.7 & 72.3 & 46.9 & 42.0 & 41.3 & 54.3 & \underline{30.3} & 26.0 & 58.5 & \underline{26.6} & 65.7 & 72.9 & \underline{25.1} & 34.8 & \underline{9.3} & 43.5 \\

				BN~\cite{BN_Stat} & 72.4 & 76.2 & 73.2 & 73.7 & 73.6 & 80.0 & 77.6 & 72.6 & 76.4 & 77.7 & 77.2 & 79.9 & 73.8 & 73.9 & 70.0 & 75.2 \\

				PL~\cite{PL} & 73.0 & 78.2 & 76.7 & 79.7 & 81.6 & 85.6 & 86.0 & 85.3 & 87.2 & 88.2 & 88.3 & 88.9 & 88.5 & 89.2 & 88.2 & 84.3 \\

				TENT~\cite{tent_wang2020} & 73.6 & 80.9 & 83.1 & 85.6 & 87.1 & 88.5 & 88.8 & 88.4 & 89.2 & 89.3 & 89.0 & 89.0 & 89.3 & 89.9 & 89.1 & 86.7 \\

				LAME~\cite{niid_boudiaf2022parameter} & 43.5 & 73.2 & 42.3 & 37.0 & 37.2 & 50.5 & \bf 22.5 & \underline{20.5} & 57.0 & \bf 18.6 & 65.5 & 71.5 & \bf 18.8 & 29.1 & \bf 5.6 & 39.5 \\

				CoTTA~\cite{cotta} & 79.5 & 81.4 & 83.4 & 83.6 & 83.9 & 85.0 & 84.0 & 82.8 & 84.8 & 84.8 & 84.5 & 84.7 & 84.1 & 84.4 & 82.8 & 83.6 \\
\end{tabular}
		}
	}
	\VspaceAfter
\end{table*}

\begin{table*}[htb]
	\centering
	\caption{Average classification error of the task CIFAR100 $\to$ CIFAR100-C while continually adapting to different corruptions at the highest severity 5 with correlatively sampled test stream under the proposed setup \setting.}
	\label{table:cifar100_order0}
	\VspaceBefore
	\resizebox{\textwidth}{!}
	{
		\renewcommand{\arraystretch}{0.8}
		{
			\begin{tabular}{l|ccccccccccccccc|c}

				\toprule[1.2pt]

				Time & \multicolumn{15}{l|}{$t\xrightarrow{\hspace*{18.5cm}}$}& \\ \hline

				Method & \rotatebox[origin=c]{45}{brightness} & \rotatebox[origin=c]{45}{pixelate} & \rotatebox[origin=c]{45}{gaussian} & \rotatebox[origin=c]{45}{motion} & \rotatebox[origin=c]{45}{zoom} & \rotatebox[origin=c]{45}{glass} & \rotatebox[origin=c]{45}{impulse} & \rotatebox[origin=c]{45}{jpeg} & \rotatebox[origin=c]{45}{defocus} & \rotatebox[origin=c]{45}{elastic} & \rotatebox[origin=c]{45}{shot} & \rotatebox[origin=c]{45}{frost} & \rotatebox[origin=c]{45}{snow} & \rotatebox[origin=c]{45}{fog} & \rotatebox[origin=c]{45}{contrast}  & Avg. \\

				\midrule

				Source & 29.5 & 74.7 & 73.0 & \underline{30.8} & 28.8 & 54.1 & \underline{39.4} & 41.2 & 29.3 & 37.2 & 68.0 & 45.8 & 39.5 & 50.3 & 55.1 & 46.4 \\

				BN~\cite{BN_Stat} & 46.5 & 52.0 & 58.6 & 47.4 & 47.4 & 57.6 & 58.2 & 56.9 & 47.0 & 53.4 & 56.0 & 52.5 & 53.1 & 57.7 & \underline{49.1} & 52.9 \\

				PL~\cite{PL} & 48.5 & 60.7 & 77.1 & 85.9 & 91.5 & 95.5 & 95.8 & 96.6 & 96.8 & 96.9 & 97.3 & 97.5 & 97.6 & 97.7 & 97.9 & 88.9 \\

				TENT~\cite{tent_wang2020} & 49.8 & 69.4 & 92.2 & 96.0 & 96.7 & 97.3 & 97.5 & 97.9 & 97.5 & 97.9 & 98.0 & 98.2 & 98.2 & 98.2 & 98.2 & 92.2 \\

				LAME~\cite{niid_boudiaf2022parameter} & \bf 21.7 & 75.1 & 72.7 & \bf 22.9 & \bf 20.6 & \underline{49.0} & \bf 32.1 & \bf 33.3 & \bf 21.2 & \bf 28.0 & 66.8 & \underline{40.0} & \bf 30.6 & \underline{43.9} & 51.3 & \underline{40.6} \\

				CoTTA~\cite{cotta} & 46.8 & \underline{48.4} & \underline{54.7} & 48.7 & 48.6 & 53.5 & 55.4 & 52.8 & 49.8 & 51.8 & \underline{53.5} & 52.9 & 54.1 & 56.7 & 53.6 & 52.1 \\
\end{tabular}
		}
	}
	\VspaceAfter
\end{table*}

\begin{table*}[htb]
	\centering
	\caption{Average classification error of the task CIFAR100 $\to$ CIFAR100-C while continually adapting to different corruptions at the highest severity 5 with correlatively sampled test stream under the proposed setup \setting.}
	\label{table:cifar100_order1}
	\VspaceBefore
	\resizebox{\textwidth}{!}
	{
		\renewcommand{\arraystretch}{0.8}
		{
			\begin{tabular}{l|ccccccccccccccc|c}

				\toprule[1.2pt]

				Time & \multicolumn{15}{l|}{$t\xrightarrow{\hspace*{18.5cm}}$}& \\ \hline

				Method & \rotatebox[origin=c]{45}{jpeg} & \rotatebox[origin=c]{45}{shot} & \rotatebox[origin=c]{45}{zoom} & \rotatebox[origin=c]{45}{frost} & \rotatebox[origin=c]{45}{contrast} & \rotatebox[origin=c]{45}{fog} & \rotatebox[origin=c]{45}{defocus} & \rotatebox[origin=c]{45}{elastic} & \rotatebox[origin=c]{45}{gaussian} & \rotatebox[origin=c]{45}{brightness} & \rotatebox[origin=c]{45}{glass} & \rotatebox[origin=c]{45}{impulse} & \rotatebox[origin=c]{45}{pixelate} & \rotatebox[origin=c]{45}{snow} & \rotatebox[origin=c]{45}{motion}  & Avg. \\

				\midrule

				Source & \underline{41.2} & 68.0 & \underline{28.8} & 45.8 & 55.1 & 50.3 & 29.3 & 37.2 & 73.0 & 29.5 & 54.1 & \underline{39.4} & 74.7 & 39.5 & 30.8 & 46.4 \\

				BN~\cite{BN_Stat} & 58.3 & 56.8 & 47.8 & 51.8 & 48.9 & 57.3 & 46.8 & 53.5 & 57.8 & 45.5 & 57.1 & 58.5 & 51.7 & 53.3 & 48.8 & 52.9 \\

				PL~\cite{PL} & 59.4 & 66.3 & 74.9 & 87.5 & 94.2 & 95.5 & 96.2 & 97.1 & 97.4 & 97.2 & 97.5 & 97.7 & 98.0 & 98.2 & 98.2 & 90.4 \\

				TENT~\cite{tent_wang2020} & 62.0 & 79.3 & 91.7 & 95.8 & 96.9 & 97.0 & 97.4 & 97.7 & 97.6 & 97.7 & 97.9 & 97.9 & 98.0 & 97.9 & 97.9 & 93.5 \\

				LAME~\cite{niid_boudiaf2022parameter} & \bf 33.6 & 66.7 & \bf 21.1 & \underline{39.9} & 50.6 & \underline{43.9} & \bf 21.0 & \bf 28.6 & 72.5 & \bf 21.6 & \underline{48.6} & \bf 32.5 & 74.5 & \bf 30.6 & \bf 22.5 & \underline{40.6} \\

				CoTTA~\cite{cotta} & 54.6 & \underline{54.1} & 49.6 & 52.1 & 52.7 & 58.0 & 50.3 & 53.3 & \underline{55.0} & 49.1 & 55.4 & 55.7 & \underline{51.0} & 54.6 & 52.1 & 53.2 \\
\end{tabular}
		}
	}
	\VspaceAfter
\end{table*}

\begin{table*}[htb]
	\centering
	\caption{Average classification error of the task CIFAR100 $\to$ CIFAR100-C while continually adapting to different corruptions at the highest severity 5 with correlatively sampled test stream under the proposed setup \setting.}
	\label{table:cifar100_order2}
	\VspaceBefore
	\resizebox{\textwidth}{!}
	{
		\renewcommand{\arraystretch}{0.8}
		{
			\begin{tabular}{l|ccccccccccccccc|c}

				\toprule[1.2pt]

				Time & \multicolumn{15}{l|}{$t\xrightarrow{\hspace*{18.5cm}}$}& \\ \hline

				Method & \rotatebox[origin=c]{45}{contrast} & \rotatebox[origin=c]{45}{defocus} & \rotatebox[origin=c]{45}{gaussian} & \rotatebox[origin=c]{45}{shot} & \rotatebox[origin=c]{45}{snow} & \rotatebox[origin=c]{45}{frost} & \rotatebox[origin=c]{45}{glass} & \rotatebox[origin=c]{45}{zoom} & \rotatebox[origin=c]{45}{elastic} & \rotatebox[origin=c]{45}{jpeg} & \rotatebox[origin=c]{45}{pixelate} & \rotatebox[origin=c]{45}{brightness} & \rotatebox[origin=c]{45}{impulse} & \rotatebox[origin=c]{45}{motion} & \rotatebox[origin=c]{45}{fog}  & Avg. \\

				\midrule

				Source & 55.1 & \underline{29.3} & 73.0 & 68.0 & 39.5 & 45.8 & 54.1 & 28.8 & 37.2 & 41.2 & 74.7 & 29.5 & \underline{39.4} & \underline{30.8} & 50.3 & 46.4 \\

				BN~\cite{BN_Stat} & 49.4 & 47.2 & 58.6 & 56.2 & 52.7 & 52.0 & 57.9 & 46.1 & 54.4 & 57.7 & 50.5 & 46.2 & 58.2 & 47.6 & 58.5 & 52.9 \\

				PL~\cite{PL} & 54.8 & 64.2 & 83.3 & 92.4 & 95.5 & 96.5 & 96.9 & 96.4 & 97.2 & 97.4 & 97.8 & 97.8 & 97.9 & 97.7 & 98.0 & 90.9 \\

				TENT~\cite{tent_wang2020} & 60.2 & 83.1 & 95.2 & 96.5 & 96.9 & 97.3 & 97.0 & 97.3 & 97.8 & 97.8 & 97.6 & 97.9 & 97.8 & 97.9 & 98.1 & 93.9 \\

				LAME~\cite{niid_boudiaf2022parameter} & 51.3 & \bf 21.3 & 72.7 & 66.3 & \bf 30.2 & \underline{40.0} & \underline{48.6} & \bf 20.9 & \bf 27.7 & \bf 33.3 & 75.0 & \bf 21.5 & \bf 32.2 & \bf 22.5 & \underline{43.8} & \underline{40.5} \\

				CoTTA~\cite{cotta} & 52.1 & 48.6 & \underline{55.1} & \underline{52.7} & 53.4 & 51.9 & 55.9 & 49.2 & 53.2 & 52.8 & \underline{49.2} & 49.7 & 56.2 & 50.7 & 58.1 & 52.6 \\
\end{tabular}
		}
	}
	\VspaceAfter
\end{table*}

\begin{table*}[htb]
	\centering
	\caption{Average classification error of the task CIFAR100 $\to$ CIFAR100-C while continually adapting to different corruptions at the highest severity 5 with correlatively sampled test stream under the proposed setup \setting.}
	\label{table:cifar100_order3}
	\VspaceBefore
	\resizebox{\textwidth}{!}
	{
		\renewcommand{\arraystretch}{0.8}
		{
			\begin{tabular}{l|ccccccccccccccc|c}

				\toprule[1.2pt]

				Time & \multicolumn{15}{l|}{$t\xrightarrow{\hspace*{18.5cm}}$}& \\ \hline

				Method & \rotatebox[origin=c]{45}{shot} & \rotatebox[origin=c]{45}{fog} & \rotatebox[origin=c]{45}{glass} & \rotatebox[origin=c]{45}{pixelate} & \rotatebox[origin=c]{45}{snow} & \rotatebox[origin=c]{45}{elastic} & \rotatebox[origin=c]{45}{brightness} & \rotatebox[origin=c]{45}{impulse} & \rotatebox[origin=c]{45}{defocus} & \rotatebox[origin=c]{45}{frost} & \rotatebox[origin=c]{45}{contrast} & \rotatebox[origin=c]{45}{gaussian} & \rotatebox[origin=c]{45}{motion} & \rotatebox[origin=c]{45}{jpeg} & \rotatebox[origin=c]{45}{zoom}  & Avg. \\

				\midrule

				Source & 68.0 & 50.3 & 54.1 & 74.7 & 39.5 & 37.2 & 29.5 & \underline{39.4} & \underline{29.3} & 45.8 & 55.1 & 73.0 & \underline{30.8} & 41.2 & 28.8 & 46.4 \\

				BN~\cite{BN_Stat} & 57.5 & 58.6 & 58.5 & 50.5 & 52.7 & 53.1 & 45.9 & 57.9 & 47.0 & 51.5 & \underline{47.8} & 58.2 & 48.2 & 57.1 & 47.7 & 52.8 \\

				PL~\cite{PL} & 59.5 & 72.9 & 85.1 & 89.6 & 94.5 & 96.8 & 97.1 & 97.9 & 97.8 & 98.0 & 98.3 & 98.2 & 98.0 & 98.0 & 98.2 & 92.0 \\

				TENT~\cite{tent_wang2020} & 60.3 & 81.4 & 95.0 & 96.6 & 97.0 & 97.3 & 97.3 & 97.7 & 97.7 & 97.7 & 97.8 & 97.7 & 97.6 & 97.6 & 97.9 & 93.8 \\

				LAME~\cite{niid_boudiaf2022parameter} & 66.4 & \bf 43.2 & \underline{49.0} & 75.2 & \bf 30.2 & \bf 28.5 & \bf 21.6 & \bf 32.5 & \bf 21.2 & \underline{39.5} & 52.0 & 72.8 & \bf 22.3 & \bf 33.1 & \bf 20.5 & \underline{40.5} \\

				CoTTA~\cite{cotta} & \underline{54.5} & 58.4 & 55.6 & \underline{50.0} & 53.9 & 53.4 & 50.3 & 56.7 & 51.3 & 53.2 & 53.7 & \underline{56.1} & 52.0 & 54.5 & 51.5 & 53.7 \\
\end{tabular}
		}
	}
	\VspaceAfter
\end{table*}

\begin{table*}[htb]
	\centering
	\caption{Average classification error of the task CIFAR100 $\to$ CIFAR100-C while continually adapting to different corruptions at the highest severity 5 with correlatively sampled test stream under the proposed setup \setting.}
	\label{table:cifar100_order4}
	\VspaceBefore
	\resizebox{\textwidth}{!}
	{
		\renewcommand{\arraystretch}{0.8}
		{
			\begin{tabular}{l|ccccccccccccccc|c}

				\toprule[1.2pt]

				Time & \multicolumn{15}{l|}{$t\xrightarrow{\hspace*{18.5cm}}$}& \\ \hline

				Method & \rotatebox[origin=c]{45}{pixelate} & \rotatebox[origin=c]{45}{glass} & \rotatebox[origin=c]{45}{zoom} & \rotatebox[origin=c]{45}{snow} & \rotatebox[origin=c]{45}{fog} & \rotatebox[origin=c]{45}{impulse} & \rotatebox[origin=c]{45}{brightness} & \rotatebox[origin=c]{45}{motion} & \rotatebox[origin=c]{45}{frost} & \rotatebox[origin=c]{45}{jpeg} & \rotatebox[origin=c]{45}{gaussian} & \rotatebox[origin=c]{45}{shot} & \rotatebox[origin=c]{45}{contrast} & \rotatebox[origin=c]{45}{defocus} & \rotatebox[origin=c]{45}{elastic}  & Avg. \\

				\midrule

				Source & 74.7 & 54.1 & \underline{28.8} & 39.5 & 50.3 & \underline{39.4} & 29.5 & 30.8 & 45.8 & 41.2 & 73.0 & 68.0 & 55.1 & 29.3 & 37.2 & 46.4 \\

				BN~\cite{BN_Stat} & 51.7 & 58.6 & 47.8 & 52.9 & 57.1 & 58.2 & 45.9 & 47.6 & 52.9 & 57.8 & 57.5 & 56.7 & \underline{49.5} & 46.1 & 54.0 & 52.9 \\

				PL~\cite{PL} & 52.4 & 68.0 & 73.4 & 87.9 & 93.7 & 96.1 & 95.7 & 96.0 & 96.5 & 96.7 & 97.5 & 97.7 & 97.7 & 97.3 & 97.7 & 89.6 \\

				TENT~\cite{tent_wang2020} & 53.5 & 77.8 & 91.1 & 96.0 & 97.0 & 97.6 & 97.4 & 97.6 & 97.9 & 98.1 & 98.1 & 98.0 & 98.1 & 97.9 & 98.1 & 92.9 \\

				LAME~\cite{niid_boudiaf2022parameter} & 74.8 & \underline{48.2} & \bf 21.1 & \bf 30.6 & \underline{43.4} & \bf 32.5 & \bf 21.6 & \bf 23.0 & \underline{39.6} & \bf 33.3 & 72.7 & 66.5 & 51.5 & \bf 20.7 & \bf 27.5 & \underline{40.5} \\

				CoTTA~\cite{cotta} & \underline{49.3} & 55.1 & 49.1 & 52.9 & 56.8 & 55.7 & 49.5 & 50.0 & 53.6 & 53.4 & \underline{54.9} & \underline{53.9} & 53.8 & 50.1 & 53.5 & 52.8 \\
\end{tabular}
		}
	}
	\VspaceAfter
\end{table*}

\begin{table*}[htb]
	\centering
	\caption{Average classification error of the task CIFAR100 $\to$ CIFAR100-C while continually adapting to different corruptions at the highest severity 5 with correlatively sampled test stream under the proposed setup \setting.}
	\label{table:cifar100_order5}
	\VspaceBefore
	\resizebox{\textwidth}{!}
	{
		\renewcommand{\arraystretch}{0.8}
		{
			\begin{tabular}{l|ccccccccccccccc|c}

				\toprule[1.2pt]

				Time & \multicolumn{15}{l|}{$t\xrightarrow{\hspace*{18.5cm}}$}& \\ \hline

				Method & \rotatebox[origin=c]{45}{motion} & \rotatebox[origin=c]{45}{snow} & \rotatebox[origin=c]{45}{fog} & \rotatebox[origin=c]{45}{shot} & \rotatebox[origin=c]{45}{defocus} & \rotatebox[origin=c]{45}{contrast} & \rotatebox[origin=c]{45}{zoom} & \rotatebox[origin=c]{45}{brightness} & \rotatebox[origin=c]{45}{frost} & \rotatebox[origin=c]{45}{elastic} & \rotatebox[origin=c]{45}{glass} & \rotatebox[origin=c]{45}{gaussian} & \rotatebox[origin=c]{45}{pixelate} & \rotatebox[origin=c]{45}{jpeg} & \rotatebox[origin=c]{45}{impulse}  & Avg. \\

				\midrule

				Source & \underline{30.8} & 39.5 & 50.3 & 68.0 & \underline{29.3} & 55.1 & 28.8 & 29.5 & 45.8 & 37.2 & 54.1 & 73.0 & 74.7 & 41.2 & \underline{39.4} & 46.4 \\

				BN~\cite{BN_Stat} & 48.5 & 54.0 & 58.9 & 56.2 & 46.4 & \underline{48.0} & 47.0 & 45.4 & 52.9 & 53.4 & 57.1 & 58.2 & 51.7 & 57.1 & 58.8 & 52.9 \\

				PL~\cite{PL} & 50.6 & 62.1 & 73.9 & 87.8 & 90.8 & 96.0 & 94.8 & 96.4 & 97.4 & 97.2 & 97.4 & 97.4 & 97.3 & 97.4 & 97.4 & 88.9 \\

				TENT~\cite{tent_wang2020} & 53.3 & 77.6 & 93.0 & 96.5 & 96.7 & 97.5 & 97.1 & 97.5 & 97.3 & 97.2 & 97.1 & 97.7 & 97.6 & 98.0 & 98.3 & 92.8 \\

				LAME~\cite{niid_boudiaf2022parameter} & \bf 22.4 & \bf 30.4 & \underline{43.9} & 66.3 & \bf 21.3 & 51.7 & \bf 20.6 & \bf 21.8 & \underline{39.6} & \bf 28.0 & \underline{48.7} & 72.8 & 74.6 & \bf 33.1 & \bf 32.3 & \underline{40.5} \\

				CoTTA~\cite{cotta} & 49.2 & 52.7 & 56.8 & \underline{53.0} & 48.7 & 51.7 & 49.4 & 48.7 & 52.5 & 52.2 & 54.3 & \underline{54.9} & \underline{49.6} & 53.4 & 56.2 & 52.2 \\
\end{tabular}
		}
	}
	\VspaceAfter
\end{table*}

\begin{table*}[htb]
	\centering
	\caption{Average classification error of the task CIFAR100 $\to$ CIFAR100-C while continually adapting to different corruptions at the highest severity 5 with correlatively sampled test stream under the proposed setup \setting.}
	\label{table:cifar100_order6}
	\VspaceBefore
	\resizebox{\textwidth}{!}
	{
		\renewcommand{\arraystretch}{0.8}
		{
			\begin{tabular}{l|ccccccccccccccc|c}

				\toprule[1.2pt]

				Time & \multicolumn{15}{l|}{$t\xrightarrow{\hspace*{18.5cm}}$}& \\ \hline

				Method & \rotatebox[origin=c]{45}{frost} & \rotatebox[origin=c]{45}{impulse} & \rotatebox[origin=c]{45}{jpeg} & \rotatebox[origin=c]{45}{contrast} & \rotatebox[origin=c]{45}{zoom} & \rotatebox[origin=c]{45}{glass} & \rotatebox[origin=c]{45}{pixelate} & \rotatebox[origin=c]{45}{snow} & \rotatebox[origin=c]{45}{defocus} & \rotatebox[origin=c]{45}{motion} & \rotatebox[origin=c]{45}{brightness} & \rotatebox[origin=c]{45}{elastic} & \rotatebox[origin=c]{45}{shot} & \rotatebox[origin=c]{45}{fog} & \rotatebox[origin=c]{45}{gaussian}  & Avg. \\

				\midrule

				Source & 45.8 & \underline{39.4} & \underline{41.2} & 55.1 & \underline{28.8} & 54.1 & 74.7 & 39.5 & 29.3 & 30.8 & 29.5 & 37.2 & 68.0 & 50.3 & 73.0 & 46.4 \\

				BN~\cite{BN_Stat} & 52.9 & 58.8 & 57.6 & 48.2 & 47.4 & 57.6 & 50.9 & 52.4 & 47.0 & 47.2 & 45.1 & 54.0 & 56.4 & 57.7 & 58.2 & 52.8 \\

				PL~\cite{PL} & 56.9 & 73.3 & 86.7 & 94.4 & 95.8 & 97.3 & 97.2 & 97.4 & 97.6 & 97.4 & 97.7 & 97.6 & 97.8 & 98.3 & 98.1 & 92.2 \\

				TENT~\cite{tent_wang2020} & 60.1 & 84.2 & 95.7 & 97.2 & 97.4 & 97.9 & 97.8 & 98.0 & 98.1 & 98.2 & 98.3 & 98.4 & 98.4 & 98.4 & 98.4 & 94.4 \\

				LAME~\cite{niid_boudiaf2022parameter} & \bf 39.9 & \bf 32.4 & \bf 33.4 & 51.4 & \bf 20.6 & \underline{49.0} & 74.4 & \bf 31.3 & \bf 21.2 & \bf 22.6 & \bf 21.9 & \bf 28.1 & 66.9 & \underline{43.9} & 72.5 & \underline{40.6} \\

				CoTTA~\cite{cotta} & 51.5 & 55.3 & 54.3 & 51.8 & 49.4 & 55.3 & \underline{50.7} & 54.2 & 51.4 & 50.6 & 49.5 & 53.6 & \underline{55.0} & 57.1 & \underline{55.8} & 53.0 \\
\end{tabular}
		}
	}
	\VspaceAfter
\end{table*}

\begin{table*}[htb]
	\centering
	\caption{Average classification error of the task CIFAR100 $\to$ CIFAR100-C while continually adapting to different corruptions at the highest severity 5 with correlatively sampled test stream under the proposed setup \setting.}
	\label{table:cifar100_order7}
	\VspaceBefore
	\resizebox{\textwidth}{!}
	{
		\renewcommand{\arraystretch}{0.8}
		{
			\begin{tabular}{l|ccccccccccccccc|c}

				\toprule[1.2pt]

				Time & \multicolumn{15}{l|}{$t\xrightarrow{\hspace*{18.5cm}}$}& \\ \hline

				Method & \rotatebox[origin=c]{45}{defocus} & \rotatebox[origin=c]{45}{motion} & \rotatebox[origin=c]{45}{zoom} & \rotatebox[origin=c]{45}{shot} & \rotatebox[origin=c]{45}{gaussian} & \rotatebox[origin=c]{45}{glass} & \rotatebox[origin=c]{45}{jpeg} & \rotatebox[origin=c]{45}{fog} & \rotatebox[origin=c]{45}{contrast} & \rotatebox[origin=c]{45}{pixelate} & \rotatebox[origin=c]{45}{frost} & \rotatebox[origin=c]{45}{snow} & \rotatebox[origin=c]{45}{brightness} & \rotatebox[origin=c]{45}{elastic} & \rotatebox[origin=c]{45}{impulse}  & Avg. \\

				\midrule

				Source & \underline{29.3} & \underline{30.8} & \underline{28.8} & 68.0 & 73.0 & 54.1 & 41.2 & 50.3 & 55.1 & 74.7 & 45.8 & 39.5 & 29.5 & 37.2 & \underline{39.4} & 46.4 \\

				BN~\cite{BN_Stat} & 47.1 & 48.6 & 47.8 & 56.2 & 57.6 & 57.6 & 57.6 & 57.5 & \underline{48.7} & 50.6 & 51.8 & 53.2 & 46.9 & 53.5 & 58.8 & 52.9 \\

				PL~\cite{PL} & 48.8 & 58.7 & 69.9 & 88.0 & 95.1 & 96.6 & 96.7 & 96.9 & 97.4 & 97.4 & 98.2 & 98.2 & 98.2 & 98.3 & 98.5 & 89.1 \\

				TENT~\cite{tent_wang2020} & 51.0 & 67.6 & 85.8 & 95.9 & 97.2 & 97.5 & 97.2 & 97.7 & 98.1 & 97.9 & 97.7 & 97.7 & 98.0 & 98.0 & 98.2 & 91.7 \\

				LAME~\cite{niid_boudiaf2022parameter} & \bf 21.2 & \bf 22.8 & \bf 21.1 & 66.3 & 72.8 & \underline{49.0} & \bf 33.3 & \underline{44.8} & 51.7 & 74.9 & \underline{39.8} & \bf 31.2 & \bf 21.3 & \bf 27.3 & \bf 32.3 & \underline{40.6} \\

				CoTTA~\cite{cotta} & 48.4 & 48.8 & 48.2 & \underline{52.9} & \underline{54.0} & 53.8 & 52.7 & 57.2 & 52.6 & \underline{48.6} & 51.8 & 53.9 & 49.4 & 52.3 & 56.0 & 52.0 \\
\end{tabular}
		}
	}
	\VspaceAfter
\end{table*}

\begin{table*}[htb]
	\centering
	\caption{Average classification error of the task CIFAR100 $\to$ CIFAR100-C while continually adapting to different corruptions at the highest severity 5 with correlatively sampled test stream under the proposed setup \setting.}
	\label{table:cifar100_order8}
	\VspaceBefore
	\resizebox{\textwidth}{!}
	{
		\renewcommand{\arraystretch}{0.8}
		{
			\begin{tabular}{l|ccccccccccccccc|c}

				\toprule[1.2pt]

				Time & \multicolumn{15}{l|}{$t\xrightarrow{\hspace*{18.5cm}}$}& \\ \hline

				Method & \rotatebox[origin=c]{45}{glass} & \rotatebox[origin=c]{45}{zoom} & \rotatebox[origin=c]{45}{impulse} & \rotatebox[origin=c]{45}{fog} & \rotatebox[origin=c]{45}{snow} & \rotatebox[origin=c]{45}{jpeg} & \rotatebox[origin=c]{45}{gaussian} & \rotatebox[origin=c]{45}{frost} & \rotatebox[origin=c]{45}{shot} & \rotatebox[origin=c]{45}{brightness} & \rotatebox[origin=c]{45}{contrast} & \rotatebox[origin=c]{45}{motion} & \rotatebox[origin=c]{45}{pixelate} & \rotatebox[origin=c]{45}{defocus} & \rotatebox[origin=c]{45}{elastic}  & Avg. \\

				\midrule

				Source & 54.1 & \underline{28.8} & \underline{39.4} & 50.3 & 39.5 & \underline{41.2} & 73.0 & 45.8 & 68.0 & 29.5 & 55.1 & 30.8 & 74.7 & 29.3 & 37.2 & 46.4 \\

				BN~\cite{BN_Stat} & 58.8 & 47.7 & 59.2 & 57.6 & 52.7 & 56.9 & 58.2 & 52.0 & 56.7 & 45.5 & \underline{47.8} & 48.2 & 51.7 & 46.1 & 54.0 & 52.9 \\

				PL~\cite{PL} & 60.1 & 59.5 & 75.1 & 85.7 & 91.5 & 94.6 & 96.5 & 97.1 & 97.4 & 97.3 & 98.0 & 97.7 & 97.9 & 97.8 & 97.7 & 89.6 \\

				TENT~\cite{tent_wang2020} & 61.6 & 71.5 & 91.0 & 95.9 & 96.6 & 97.1 & 96.9 & 97.3 & 97.4 & 97.2 & 97.9 & 98.0 & 98.1 & 97.9 & 97.8 & 92.8 \\

				LAME~\cite{niid_boudiaf2022parameter} & \underline{48.6} & \bf 20.6 & \bf 32.3 & \underline{44.4} & \bf 30.2 & \bf 33.6 & 72.4 & \underline{40.0} & 66.3 & \bf 21.6 & 52.0 & \bf 22.8 & 74.6 & \bf 20.7 & \bf 27.5 & \underline{40.5} \\

				CoTTA~\cite{cotta} & 56.4 & 48.9 & 56.1 & 57.8 & 54.1 & 54.2 & \underline{56.2} & 53.6 & \underline{55.4} & 50.0 & 53.6 & 51.6 & \underline{51.2} & 50.7 & 54.4 & 53.6 \\
\end{tabular}
		}
	}
	\VspaceAfter
\end{table*}

\begin{table*}[htb]
	\centering
	\caption{Average classification error of the task CIFAR100 $\to$ CIFAR100-C while continually adapting to different corruptions at the highest severity 5 with correlatively sampled test stream under the proposed setup \setting.}
	\label{table:cifar100_order9}
	\VspaceBefore
	\resizebox{\textwidth}{!}
	{
		\renewcommand{\arraystretch}{0.8}
		{
			\begin{tabular}{l|ccccccccccccccc|c}

				\toprule[1.2pt]

				Time & \multicolumn{15}{l|}{$t\xrightarrow{\hspace*{18.5cm}}$}& \\ \hline

				Method & \rotatebox[origin=c]{45}{contrast} & \rotatebox[origin=c]{45}{gaussian} & \rotatebox[origin=c]{45}{defocus} & \rotatebox[origin=c]{45}{zoom} & \rotatebox[origin=c]{45}{frost} & \rotatebox[origin=c]{45}{glass} & \rotatebox[origin=c]{45}{jpeg} & \rotatebox[origin=c]{45}{fog} & \rotatebox[origin=c]{45}{pixelate} & \rotatebox[origin=c]{45}{elastic} & \rotatebox[origin=c]{45}{shot} & \rotatebox[origin=c]{45}{impulse} & \rotatebox[origin=c]{45}{snow} & \rotatebox[origin=c]{45}{motion} & \rotatebox[origin=c]{45}{brightness}  & Avg. \\

				\midrule

				Source & 55.1 & 73.0 & \underline{29.3} & 28.8 & 45.8 & 54.1 & 41.2 & 50.3 & 74.7 & 37.2 & 68.0 & \underline{39.4} & 39.5 & 30.8 & 29.5 & 46.4 \\

				BN~\cite{BN_Stat} & 49.5 & 58.8 & 47.0 & 46.5 & 52.2 & 57.6 & 57.6 & 57.6 & 51.7 & 53.5 & 56.0 & 58.5 & 53.1 & 47.6 & 46.3 & 52.9 \\

				PL~\cite{PL} & 53.6 & 70.4 & 76.0 & 85.1 & 91.2 & 95.2 & 96.0 & 97.0 & 96.9 & 97.3 & 97.3 & 97.6 & 97.5 & 97.6 & 97.7 & 89.8 \\

				TENT~\cite{tent_wang2020} & 60.2 & 89.1 & 95.0 & 96.2 & 96.9 & 97.0 & 96.5 & 97.0 & 97.0 & 97.2 & 97.6 & 97.8 & 97.5 & 97.9 & 97.7 & 94.0 \\

				LAME~\cite{niid_boudiaf2022parameter} & 51.3 & 72.5 & \bf 21.5 & \bf 21.0 & \underline{39.6} & \underline{49.0} & \bf 33.3 & \underline{44.8} & 74.8 & \bf 28.0 & 66.8 & \bf 32.5 & \bf 30.6 & \bf 22.5 & \bf 21.4 & \underline{40.6} \\

				CoTTA~\cite{cotta} & 52.3 & \underline{55.3} & 49.5 & 48.1 & 52.1 & 54.8 & 52.7 & 56.9 & \underline{50.6} & 52.6 & \underline{53.7} & 55.8 & 54.6 & 50.6 & 50.5 & 52.7 \\
\end{tabular}
		}
	}
	\VspaceAfter
\end{table*}

\end{document}